\documentclass{article}
\usepackage[english]{babel}

\usepackage[letterpaper,top=2cm,bottom=2cm,left=3cm,right=3cm,marginparwidth=1.75cm]{geometry}

\usepackage{graphicx}
\usepackage[colorlinks=true, allcolors=blue]{hyperref}
\usepackage{subcaption}

\usepackage[authoryear]{natbib}

\usepackage{hyperref}
\hypersetup
{
 unicode=false,     
 pdftoolbar=true,    
 pdfmenubar=true,    
 pdffitwindow=false,   
 pdfstartview={FitH},  
 pdftitle={My title},  
 pdfauthor={Author},   
 pdfsubject={Subject},  
 pdfcreator={Creator},  
 pdfproducer={Producer}, 
 pdfkeywords={keywords}, 
 pdfnewwindow=true,   
 colorlinks=true,    
 linkcolor=red,     
 citecolor=blue,    
 filecolor=magenta,   
 urlcolor=blue      
}

\usepackage[figurename=Fig.,labelfont=bf,labelsep=period]{caption}
\usepackage{subcaption}
\usepackage{amsmath}
\usepackage{amsfonts}
\usepackage{amssymb}
\usepackage{amsbsy}
\usepackage{booktabs}
\usepackage{colortbl}
\usepackage[dvipsnames]{xcolor}
\usepackage[none]{hyphenat}
\usepackage{pdflscape}
\usepackage{multirow}
\usepackage{array}
\usepackage{pdfpages}

\usepackage{lmodern}

\usepackage[T1]{fontenc}
\usepackage[final, babel]{microtype} 

\usepackage{lipsum}         

\usepackage{lscape}

\title{Predicting travel demand of a bike sharing system using graph convolutional neural networks}

\author{
Ali Behroozi \footnote{School of Civil Engineering, KNTU, Iran. Email: Ali.Behroozi@email.kntu.ac.ir}
\and
Ali Edrisi \footnote{Corresponding author. School of Civil Engineering, KNTU, Iran. Email: edrisi@kntu.ac.ir}
}

\begin{document}

\maketitle
\newpage

\begin{abstract}

Public transportation systems play a crucial role in daily commutes, business operations, and leisure activities, emphasizing the need for effective management to meet public demands. One approach to achieve this goal is by predicting demand at the station level. Bike-sharing systems, as a form of transit service, contribute to the reduction of air and noise pollution, as well as traffic congestion. This study focuses on predicting travel demand within a bike-sharing system. A novel hybrid deep learning model called the gate graph convolutional neural network is introduced. This model enables prediction of the travel demand at station level. By integrating trajectory data, weather data, access data, and leveraging gate graph convolution networks, the accuracy of travel demand forecasting is significantly improved. Chicago City bike-sharing system is chosen as the case study. In this investigation, the proposed model is compared to the base models used in previous literature to evaluate their performance, demonstrating that the main model exhibits better performance than the base models. By utilizing this framework, transportation planners can make informed decisions on resource allocation and rebalancing management.

\end{abstract}


\section{Introduction}
\label{sec:intro}

Cities are getting larger due to factors such as population growth, economic opportunities, infrastructure developments, congestion effects, urbanization, and government policies. By 2050, it is forecast that 70\% of the world's population would reside in cities \citep{wang2020deep}. This progress of urbanization has engendered significant challenges in short-term demand forecasting for public transportation systems.

The emergence of novel technological advancements has provided the ability to utilize vast reserves of big data, such as information gathered from public transit cards, to identify and fully comprehend travel patterns, including demand profiles \citep{ke2018hexagon}, temporal distribution \citep{yildirimoglu2013experienced}, and spatial and other patterns analyses \citep{ma2013mining,lahoorpoor2019spatial}. Travel demand and mobility patterns are the most critical factors in transport planning and operation optimization problems, which further affect the effectiveness, dependability, and appeal of public transit services.

The bike-sharing system is a transit service that can be utilized for short commutes, occasional outings, and leisure visits. This mode of transport has shown numerous advantages at the individual, municipal, and regional levels. The production of a car emits far more CO2 and uses significantly more energy than a bicycle, and the same is true for bike trips versus automobile trips \citep{shaheen2007reducing, weiss2015electrification}. In terms of local advantages, bike trips reduce air and noise pollution, traffic congestion, and the need for parking spaces \citep{saelensminde2004cost}. In response to the COVID-19 pandemic, many cities have provided bike lanes to increase social distancing and to avoid buses and subways. In addition, bike sharing has the potential to enhance urban mobility as a sustainable mode of transportation.

One of the challenges in improving the performance of bike-sharing systems is predicting travel demand; Addressing this problem will enhance rebalancing schemes by accurately predicting travel demand, which is influenced by a range of spatial and temporal parameters. These parameters include demographics, activity locations, network topology, time of day, weather conditions, and more. Integrating all these factors into a single model demands not only comprehensive data but also advanced modeling techniques. With the availability of large data and machine learning techniques, new methods have been developed to predict travel demand including long short-term memory (LSTM), gated recurrent unit (GRU), convolutional neural network (CNN), and graph convolutional neural network (GCNN).

LSTM and GRU are powerful recurrent neural network methods for forecasting time series, particularly effective when dealing with datasets exhibiting long-term trends, as they allow for learning a greater number of parameters. Therefore, they are able to obtain temporal features and the trend of traffic flow. CNN extracts spatial correlation using regular (Euclidean) data, and each grid in the network is treated as a node, with neighbors determined by the size of the filter. It takes the weighted average of the grid values of the node and its neighbors, however, it has shortcomings in large-scale travel demand forecasting. By abstracting the network into a graph (e.g., bike-sharing stations as the vertices and bike-sharing lines as the edges), GCNN can be used to effectively capture the irregular spatiotemporal dependencies. In comparison to traditional approaches, using GCNNs for traffic prediction may provide improved accuracy and scalability.

To summarize, oversimplified schemes and models cannot accurately predict the travel demand of bike-sharing systems, and also research in the past has focused on building demand prediction models for each station separately without considering hidden correlations between stations in order to improve the performance of the model \citep{faghih2014land,rixey2013station,yang2016mobility}. To bridge this gap, this study proposes a gate graph neural network deep learning method and considers various spatiotemporal features such as weather conditions, historical demand, and access. This research assumes that the graph is changed at each time step due to the dynamic nature of traffic. The model is tested for the Chicago bike-sharing system and compared against a classic econometric model (OLS), as well as other machine learning (SVR, XGBoost), and deep learning models (GCN, MLP, LSTM, CNN). Additionally, various types of graph convolutional neural networks with attention mechanisms and integration with temporal methods are explored.

The remainder of this paper is structured as follows. A brief historical background refers to the models and parameters that have been employed in forecasting travel demand. This context provides insight into the established approaches and factors considered in previous research related to predicting travel demand in \autoref{sec:lit}. In \autoref{sec:method} definition of primary concepts such as graph, graph convolution, dynamic graph, and adjacency matrix, as well as mathematical definitions such as graph convolution, gate recurrent units, and ultimately, gate graph convolution neural networks, as well as measuring access method and definition base models, will be continued. In \autoref{sec:case}, we provide a brief overview of the case study and explain variables such as bike share transactions, weather conditions, and demographic statistics. The results of the statistical experiments are provided in \autoref{sec:result}. Finally, the concluding remarks are discussed in \autoref{sec:discussion}.

\section{Literature review}
\label{sec:lit}
The primary challenge in forecasting bike sharing lies in the intricacies associated with spatial and temporal modeling. Firstly, the distribution of demand across various locations is significantly imbalanced. Secondly, demand peaks during the morning and evening rush hours as a result of a higher volume of commuters. However, this temporal pattern is not consistently observed spatially during weekends and holidays, and dependencies on external factors like social events or weather further complicate demand prediction. Thirdly, the intricate spatial and temporal dependencies undergo changes with alterations in bike share systems or the influences of adjacent regions \citep{jiang2022bike, luo2021predicting}.

Several demand models for different transport modes have been established in the literature. These models can be categorized as econometric models, data-driven machine learning models and recently deep learning techniques to estimate the demand between any origin-destination pair.

Dominant econometric and statistical models include the classical four-step model (i.e., trip generation, trip distribution, modal split, and traffic assignment), the historical average, and the auto-regressive integrated moving average model \citep{amini2016arima, pan2012utilizing}.

Recently, investigations have shown an increase in the popularity of machine learning in the field of travel demand modeling. Recent studies have demonstrated that machine learning can significantly improve predictions of travel mode choice at the individual level when compared to traditional logit models \citep{xie2003work,hagenauer2017comparative,cheng2019applying}. Support vector regression, random forest, and gradient boosting decision trees are examples of machine learning models for travel demand forecasting; Researchers have attempted to forecast travel demand using machine learning (trip frequency rather than mode choice is the outcome variable) such as public transit \citep{ma2018parallel}, ride-sourcing \citep{geng2019spatiotemporal}, and bike-sharing \citep{lin2018predicting}. The models can learn temporal dependencies from historical demand; however, they exhibit limitations in simultaneously capturing intricate and dynamic spatial-temporal dependencies.

With recent advances in technology and parallel computing powers, transport demand problems have been addressed through the application of deep learning techniques, including neural networks and their derivative forms such as recurrent neural networks or convolutional neural networks. Recurrent Neural Networks (RNNs) are designed for time series problems and retain temporal information; however, they encounter challenges, such as the problem of vanishing gradient stemming from the use of tanh activation in the hidden state to merge previous information and current input features. Gated Recurrent Units (GRU) and Long Short-Term Memory (LSTM) models represent enhanced iterations of RNNs, addressing the issues inherent in traditional RNNs. These models incorporate a gate mechanism, facilitating the updating of historical information while mitigating computational problems. They are particularly adept at capturing long-term temporal dependencies. On the other hand, Convolutional Neural Networks (CNNs) and Graph Convolutional Networks (GCNs) are commonly employed to address spatial dependencies in data \citep{chen2020predicting, pan2019predicting, jiang2022bike, he2020towards, xiao2021demand, zhang2021traffic}.

In previous studies, it has been shown that deep learning techniques outperformed the classical statistical models and various machine learning models \citep{ke2021predicting,kim2019graph,lin2018predicting}. 
Since the graph neural network and knowledge graph have become more widely used, recent studies focus on dynamic graphs rather than just static ones. A more comprehensive methodology for dynamic graph creation involves learning how to describe dynamic networks and, for example, performing travel demand predictions on each link in networks based on the learned representations \citep{trivedi2019dyrep, goyal2018dyngem}.

The data-driven bike-sharing demand has been investigated in the literature specifically for the re-balancing problem and making the bikes available to more potential users \citep{lahoorpoor2019spatial,ma2021rebalancing,wang2022dynamic,caggiani2018modeling,feng2017analysis,dondo2007cluster,du2019model}. However, these studies focused on the optimization aspect of the problem, and little attention has been paid to the accuracy of their demand model if any. There are a few studies which implemented machine learning techniques on estimating demand between bike-sharing stations \citep{cho2021enhancing}. 

For example, \citet{ashqar2022network} developed a model for predicting bike availability in the San Francisco metropolitan area using the Random Forest and Least-Squares Boosting algorithms. In a similar case study, \citet{yan2020using} employed a random forest model to estimate Chicago's zone-to-zone demand. They have shown that random forest models outperform traditional multiplication models in terms of model fit and forecast accuracy. Multiplicative models are quite popular and have many applications, such as time series analysis and forecasting. A coefficient, also known as elasticity, shows the percent change in the dependent variable when the independent variable changes by one percent, while holding all other variables constant \citep{mohr2021statistical}.  

\citet{giot2014predicting} compared the performance of a several of machine learning methods, including Ridge Regression, Adaboost Regression, Support Vector Regression, Random Forecast Tree, and Gradient Boosting Regression Trees. They found that the Ridge Regression, Adaboost Regression performed better in predicting the bike demand. \citet{zhou2015understanding} used the Agglomerative Hierarchical Clustering technique and the Community Detection algorithm to group similar bike flows and stations in Chicago. \citet{deng2018integrating} created demand models for pick-ups and drops for the Vienna-based Citybike Wien system. \citet{rudloff2014modeling} used count models such as the Hurdle, Poisson, and Negative Binomial. They demonstrated the significance of adding new stations when modeling the demand function by taking meteorological conditions into account as influences on demand. \citet{yang2016mobility} created a spatio-temporal model to forecast station- and time-based traffic using a random forest algorithm. Pressure, humidity, visibility, wind direction, wind speed, temperature, dew point, and conditions from the last 30 minutes were combined. The evaluation results indicate that there is a 0.6 relative error at the 85th percentile for predicting both check-in and check-out events.

\citet{yu2017spatio} introduced spatial-temporal graph convolution neural network(STGCN) framework was suggested. The traffic network was modeled using a general graph structure, where the stations were depicted as nodes to fully capture spatial information. The graph's nodes used graph convolution and gated linear units to extract spatial and temporal properties, respectively, despite having varying observation values at various points in time. A method of predicting traffic flow using multi-graph convolutional networks. \citet{chai2018bike} a variety of graphs based on distances, traffic interactions, and correlations between stations in the traffic network were created and fused to learn the spatio-temporal properties of the network. \citet{liu2017short} CNN-LSTM, a model that extracts the spatiotemporal characteristics of traffic flow by mapping traffic flow data to a one-dimensional vector space, combining the one-dimensional spatial information vectors at various periods into a matrix, and convolving using CNN and LSTM, was presented.

 \citet{sankar2018dynamic} presented the Dynamic Self-Attention Network(DySAT) model, which predicted linkages by mastering the representation of a dynamic graph's nodes and using the dynamic graph's attention mechanism to capture changes in the same node at all times. \citet{trivedi2019dyrep} developed the DyRep model, which encodes these events to learn link prediction by characterizing dynamic graph changes as events. \citet{han2021dynamic} dynamic graph neural networks(DGNNs), have a technique for predicting traffic speed. This means that future traffic speeds will be determined by various factors, including traffic volumes, in addition to current speeds. Aim to explore the dynamic and complex spatio-temporal properties of traffic data to better leverage the capabilities of DGNNs for more accurate traffic speed forecasts. Create a dynamic graph creation technique specifically designed to discover the spatio-temporal relationships among road segments. Subsequently, introduce a dynamic graph convolution module that, by passing messages on dynamic adjacency matrices, aggregates hidden states of neighboring nodes to the focal nodes.

\citet{ma2022short} introduced the Spatial-Temporal Graph Attentional Long Short-Term Memory (STGA-LSTM) model, which includes an attentional mechanism to predict short-term demand at the station level. Various input features, including historical bike-sharing trip data for the month of September 2017, weather data, user information, and land-use data, were utilized. \citet{li2022data} introduced a Spatial-Temporal Graph Neural Network named STGNN-DJD, incorporating Dynamic and Joint Spatial-Temporal Dependency (ST Dependency). This network utilizes a graph generator to construct two graphs: the first one capturing flow relationships between stations, and the second modeling dynamic correlations in demand-supply patterns among stations. \citet{huang2022dynamical} mentioned that dynamic demand highlights the challenge of forecasting demand. To address this, they introduce a Dynamic Spatial-Temporal Graph Neural Network consisting of two prominent components. The first component involves creating a spatial dependence graph, which is a spatial graph based on the stability of nodes, capturing the dynamic relationship between nodes. The second component is inferring intensity, which involves modeling changing demand and eventually integrating it with a diffusion convolution neural network and a transformer.

When predicting the demand for bike-sharing, these characteristics should be taken into account because they have a significant impact on how often bikes are used. Although some of these parameters have been considered in certain previous demand forecasting models \citep{ke2021predicting, tang2021multi, yang2020station}, few studies have thoroughly incorporated these factors into their models for projecting the real-time bike-sharing demand, particularly the weather conditions. 

Several studies have extracted historical transaction data from Divvy bike-sharing. For example, \citet{zi2021tagcn} utilized historical transactions from 2019 and implemented a Temporal Attention Graph Convolution Network (TAGCN) to forecast travel demand for bike-sharing across various time spans. Additionally, \citet{wang2021modeling} proposed a regression model that considers spatially varying coefficients. They observed the effects of land use, transportation infrastructure, and socio-demographic factors on travel demand for bike-sharing. \citet{he2020towards} proposed Gbike model, an architecture that consists of multi-level graph attention networks and takes into account external features, such as weather, points-of-interest (POIs), and events.

\section{Methodology}
\label{sec:method}
In the preceding sections, we discussed the research question and the models employed in previous studies. Firstly, this section presents precise definitions of fundamental concepts such as defining a graph, a fundamental description of the graph convolutional neural network, dynamic graph, and adjacency matrix. Additionally, it introduces the concepts of graph convolutional neural network and the gate recurrent unit. Lastly, it puts forth the principal framework, namely the gate graph convolution neural network, and elaborates on the locational access. One way to investigate the performance of the model is by comparing it with the base models mentioned in the literature. In the last subsection, we describe the structure of these base models.

The nature of the data plays a decisive role in selecting a deep learning framework. For instance, when dealing with sequential data, the use of Recurrent Neural Networks (RNNs) is suitable for extracting temporal features. As highlighted in the literature review, for data characterized by a non-Euclidean space where each node has characteristics and is interconnected with other nodes, the Graph Convolutional Network (GCN) emerges as a fitting choice. GCNs excel in capturing localized patterns by focusing on the connections between neighboring nodes. Introducing an appropriate graph to the GCN can yield effective results by extracting both spatial and temporal features from the graph. A notable advantage is the ability to seamlessly integrate GCN with other deep learning frameworks, which will be elaborated upon in this study, particularly in the implementation of GCN-LSTM.

Previous research made the assumption that spatially neighboring regions could have an effect on OD demand patterns \citep{bai2019passenger,ke2018hexagon}. Nevertheless, demand also significantly relies on non-Euclidean aspects, including demographic profile (e.g., sex, age, and income) and land use characteristics (location of opportunities and point of interest) around origins and destinations. Deep learning models built on graphs can process non-Euclidean data and efficiently extract characteristics from station-level demand. 

In this paper, we define a directed graph $G=(V,E,A)$ and use a gated method for graph convolution layers, where V represents the set of stations, E is the set of constant-time edges, and A is the adjacency matrix. An edge in the graph reflects the correlation between two station pairs in each direction.

The proposed method to predict stations demand  comprises a series of gated graph convolution neural networks (GGCNN) to extract the spatial-temporal interactions between stations. The structure of the graph is defined based on the stations demand.

Weather conditions, including temperature, precipitation, and wind speed, can significantly influence the usage of bike-sharing systems \citep{mathew2019impact, noland2021scootin, younes2020comparing}. The demand is also influenced by land use type and patterns \citep{bai2020dockless, mitra2021potential, sanders2020scoot}. Therefore, we consider these features in the proposed model, which will be discussed in the following subsections.

\subsection{Defining graph convolutional neural network}

Convolutional neural network (CNN) is primarily intended to extract statistically meaningful patterns and learn local structures in a dataset, including similar grids. Recent studies have suggested using graph CNN (GCNN) for data on irregular patterns, such as road networks with varying connectivity, transportation systems with diverse structures, or geographical regions with uneven distributions \citep{bruna2013spectral,sandryhaila2013discrete,shuman2013emerging}.

\subsection{Dynamic graph construction}

Due to the dynamic nature of the traffic, implicit and dynamic spatial dependencies between segments must be taken into account while designing a graph learning module. In prior studies, dependencies between nodes were either produced as a static graph that omitted the dynamic property of traffic spatial relationships or were determined by human knowledge, such as proximity and functional similarity.\\ 
In this study, we present the dynamic graph constructor, which generates dynamic learnable graphs utilized in the dynamic gate graph convolution module to capture changing interactions between nodes. We presume that the traffic at each hour could give rise to a distinct graph compared to previous ones, owing to the variable travel demand between stations. As a result, our objective is converted to an adjacency tensor $A \in \mathbf{R^{N_txNxN}}$, where $N_t$ is the total number of daily time slots.

\subsection{Adjacency matrix construction}
An adjacency matrix indicates the spatial correlation between adjacent stations, assigning weights to connected nodes. In order to get the similarity between regions $r_i$ and $r_j$, a direct way is to calculate the correlations (e.g., the Pearson Coefficient) by the transactions for bike sharing combined by the hour.\\ By calculating the Pearson Correlation Coefficient (PCC) using the hourly bike demand series between stations i and j, the Demand Correlation Matrix is defined \citep{lin2018predicting}.

\begin{equation}
    G=(V,E,A) \hspace{1cm}  \text{weight} = PCC(h_i , h_j)
\end{equation}
\begin{figure}[!ht]
  \centering
  \includegraphics[width=10cm,height=10cm]{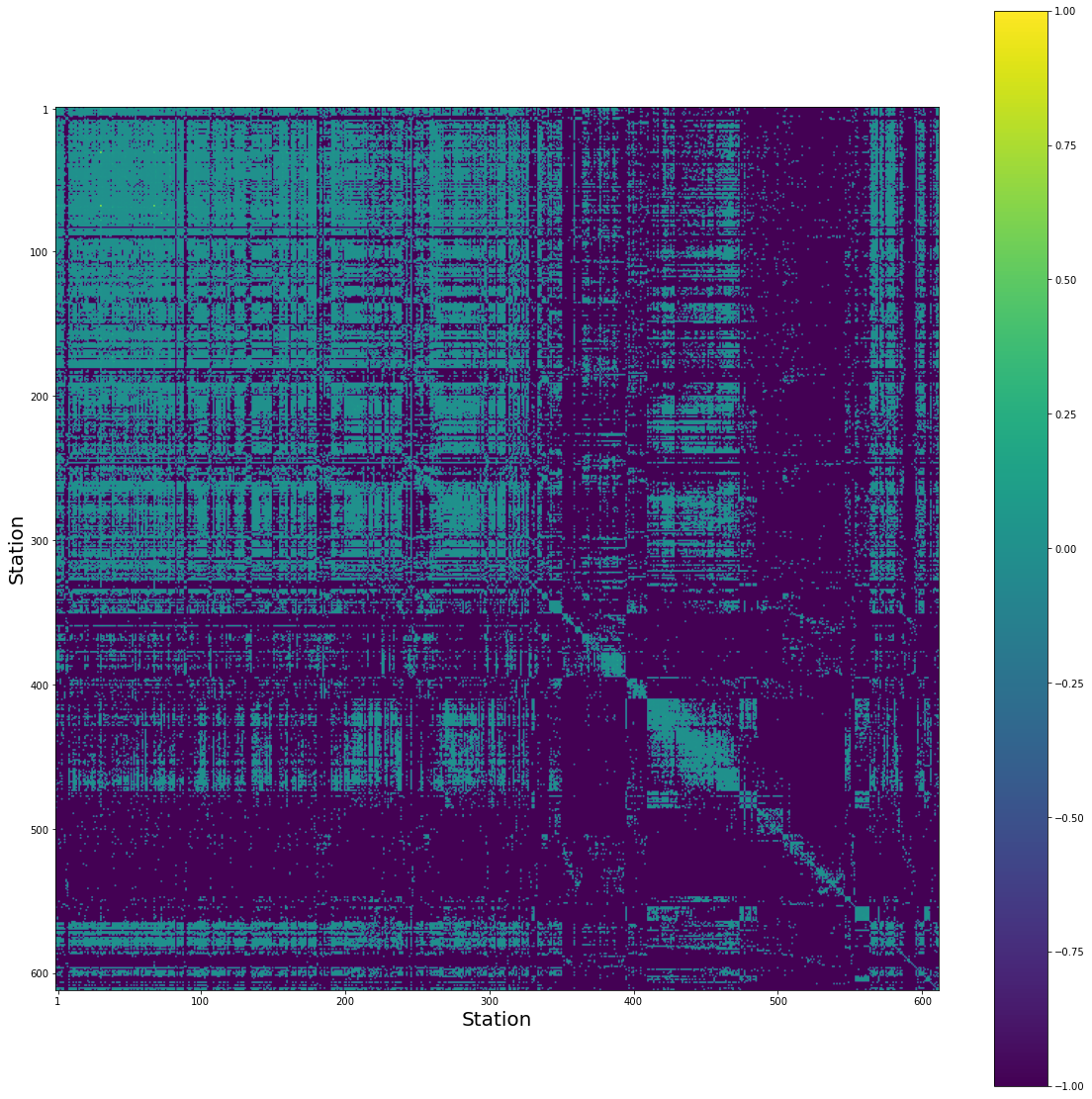}
  \caption{Correlation of travel demand between stations} \label{fig:Correlation stations}
\end{figure}

where the hourly bike demand series for stations $i$ and $j$ are $h_i$ and $h_j$. This approach builds the bike-sharing graph network by connecting two stations with quantity PCC. \autoref{fig:Correlation stations} demonstrates the correlation between stations by capturing the travel demand patterns. It highlights that each station exhibits a strong correlation with itself, indicating a higher concentration of travel demand within individual stations.

\subsection{Graph convolution neural network}
The prediction of traffic flow is a complex task that involves considering various factors such as traffic conditions, weather conditions, and access to population and employment centers (denoted as feature X) for each station in the traffic network at a given time (t). This feature X represents the inflow or demand at each station. Short-term forecast to anticipate the traffic flow $X^{t+n}$ at time step $t+n$ based on known traffic flow is a traffic flow prediction task $[X^1, \dots, X^t]$. In this paper, where the actual time period corresponding to n is smaller than a day, we define the problem as a short-term prediction \citep{peng2021dynamic}. Consider a graph G=(V,E,A), where V is a finite collection of vertices with size N, E is a set of edges, and A is the adjacency matrix. Entry $A_{ij}$ encodes the spatial correlation of connection between two vertices. The definition of a normalized graph Laplacian matrix as $L \in \mathbf{R^{NxN}}$ is:

\begin{equation}
    L=I_N - D^{-1/2} A D^{-1/2}
\end{equation}

Where $D \in \mathbf{R^{NxN}} $ a diagonal degree matrix and $I_N$ is the identity matrix $D_{ii} = \Sigma _j A_{ij}$ .
The diagonal of the real symmetric positive semi-definite matrix L is:

\begin{equation}
    L=U \Lambda U^T
\end{equation}

Where $U=[u_0, u_1, \dots, u_{N-1}]$ ; $\Lambda = \text{diag}([\lambda_0, \lambda_1, \dots, \lambda_{N-1}])$; $\Lambda \in \mathbf{R^{NxN}}$ are the eigenvalues of L, and $U \in \mathbf{R^{NxN}}$ is the associated group of orthonormal eigenvectors. \\On the graph, a spectral convolution is described as follows:

\begin{equation} \label{eq:4}
    g_\theta * x = Ug_\theta (\Lambda) U^T x
\end{equation}

Where $g_\theta (\Lambda)$ is an eigenvalue function of L.\\ In this study, Chebyshev polynomials $T_K(x)$ are used to approximate the filter function $g_\theta (\Lambda)$ \citep{defferrard2016convolutional}. Therefore, \autoref{eq:4} can be approximated using an approximation method proposed by \citet{welling2016semi}:

\begin{equation}
    x*g_\theta = \Theta(I_N + M^{-1/2} A M^{-1/2})x  
\end{equation}

In this case, M represents the diagonal matrix of node degrees, and $M_{ii}=\Sigma_j A_{ij}$. Graph convolution neural network (GCNN) output can thus be estimated as follows:

\begin{equation}\label{eq:6}   
    X^{l+1}=(\tilde{M}^{-1/2} \tilde{A} \tilde{M}^{-1/2}) X^l W  
\end{equation}  

\begin{figure}[!ht]
  \centering
  \includegraphics[width=14cm,height=10cm]{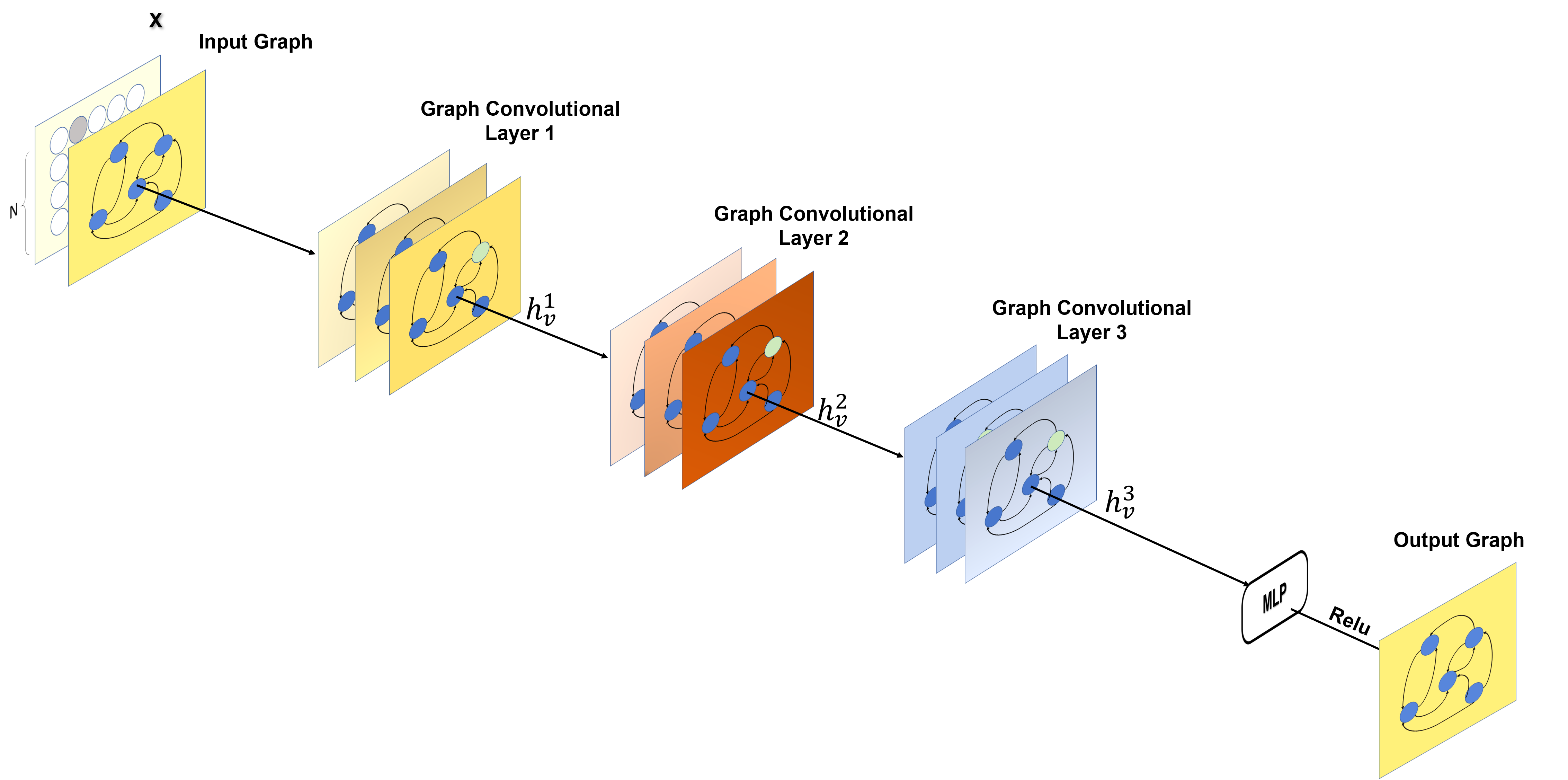}
  \caption{Graph convolution layers \\ MLP: Multi Linear Perceptron; X: Features nodes of the graph} \label{fig:GCN layer}
\end{figure}

So $I_N + M^{-1/2} A M^{-1/2}$ is replace by $\tilde{M}^{-1/2} \tilde{A} \tilde{M}^{-1/2}$, and $\tilde{A}=A+I_n$, $\tilde{M_{ii}}=\Sigma_j \tilde{A_{ij}}$. The demand-like input is represented by X in the $l^{th}$ GCNN. Notably, there are no non-linear operations in the step of \autoref{eq:6}.
\autoref{fig:GCN layer} provides a visual representation of the input layer, comprising features derived from the nodes matrix and the graph. Additionally, \autoref{fig:GGCN} demonstrates the presence of three crucial layers: graph convolutional (GCN) and linear layers (MLP), as well as an output layer, collectively responsible for predicting the graph in the next time step.

\subsection{Gated Recurrent Unit Cell}
The LSTM cell demonstrates enhanced learning capabilities compared to a conventional recurrent cell, although the additional parameters introduce computational complexity. The gated recurrent unit (GRU) has a simpler structure with fewer gates, making it easier to understand and implement. The simpler structure of GRUs than LSTM contributes to quicker training and inference time, thereby enhancing the computational efficiency of these models \citep{shiang2020gated}. The GRU cells' mathematical expressions are as follows:
\begin{equation} \label{eq:7}
    r_t=\sigma(W_{ir} x_t + b_{ir} + W_{hr} h_{t-1} + b_{hr})
\end{equation}
\begin{equation}
    z_t=\sigma(W_{iz} x_t + b_{iz} + W_{hz} h_{t-1} + b_{hz})
\end{equation}
\begin{equation}
    \tilde{h_t}=\tanh(W_{in}x_t + b_{in} + r_t * (W_{hn} h_{t-1} + b_{hn}))
\end{equation}
\begin{equation}
    h_t=(1-z_t)*\tilde{h_t} + z_t*h_{t-1}
\end{equation}

Where $h_t$ is the hidden state at time t, $x_t$ is the input at time t, $h_{t-1} \in \mathbf{R^{NxN}}$ is the hidden state of the layer at time $t-1$ or initial hidden state at time 0, and $r_t$,$z_t$,$n_t$ are the reset, update, and new gate, respectively \citep{cho2014learning}.
Where the sigmoid function, denoted by the $\sigma$, is defined as:

\begin{equation}
    \sigma (x)= \frac{1}{1+e^{-x}}
\end{equation}

\begin{figure}[!ht]
  \centering
  \includegraphics[width=13cm,height=10cm]{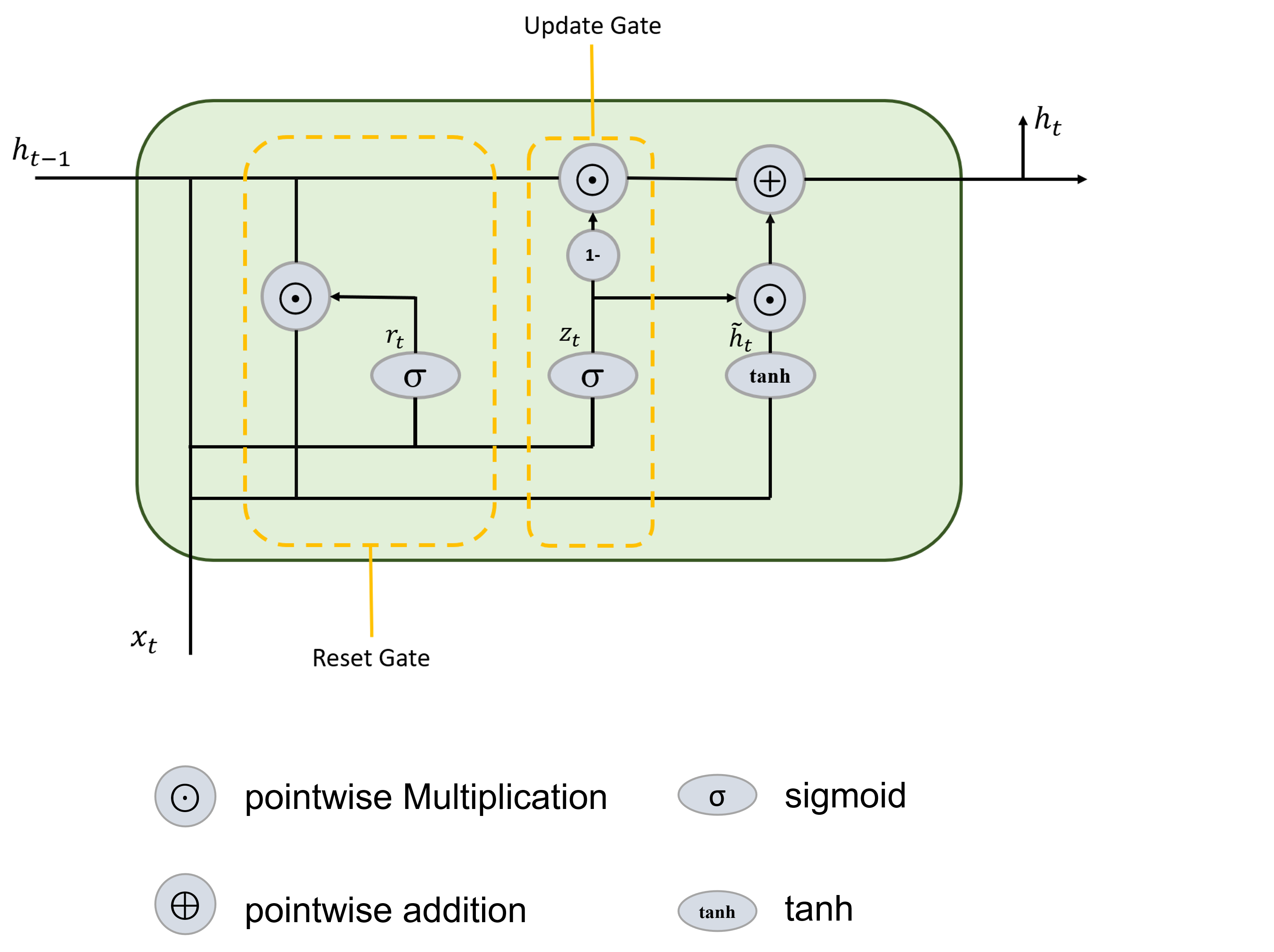}
  \caption{Gate Recurrent Unit Cell structure} \label{fig:GRU Cell}
\end{figure}

\autoref{fig:GRU Cell} depicts the structure of a GRU cell, which is composed of two essential components: the reset gate and the forget gate. The input to the cell includes features at time $t$ and the hidden state at time $t-1$. The resulting output of the GRU cell is the hidden state at time $t$, representing the accumulated information and encoding the relevant features extracted from the input sequence.

\subsection{Vanish gradient}
The vanishing gradient problem is influenced by several factors, including the activation functions employed in RNNs, such as the sigmoid or hyperbolic tangent functions. These functions tend to saturate, resulting in gradients that approach zero. Additionally, the recurrent nature of RNNs, where gradients are multiplied across multiple time steps during backpropagation, exacerbates the issue. In order to address this problem, advanced architectures like LSTM (Long Short-Term Memory) and GRU (Gated Recurrent Unit) were introduced \citep{hochreiter1997long}. These architectures incorporate gate cells, which prevent vanishing gradients. For our investigation, we have utilized gate cells in the graph convolutional model to mitigate the vanishing gradient issue.

\subsection{Framework overview}
\label{sec:Framework overview}
Recurrent neural network (RNN)-based models are frequently employed in time-series prediction. However, they face challenges related to time-intensive training and intricate gate mechanisms, particularly within a large-scale transportation network. GCNN is superior than RNN-based models in that it has a simple design and quick training. To meet the temporal dependency of the demand for public bike sharing, the gated GCNN is used in this study \citep{xiao2021demand}.

The most significant change to GCNN in this paper is the use of gate recurrent units. The GCNN uses a gating technique to account for non-linearity and determine which elements of the linear transformation can pass past the gate and so affect the prediction \citep{dauphin2017language}. Furthermore, the vanishing gradient issue is minimized by residual learning \citep{he2016deep}. These steps result in the proposed Gated Graph Convolutional neural network construction being complete (GGCNN). The outcome of $l^{th}$ layer is expressed as follows:

\begin{equation} \label{eq:12}
    h_i^{(0)}=x_i || 0
\end{equation}
\begin{equation}
    m_i^{(l+1)}=\Sigma_{j \in N(i)} A_{ji} W h_j^{l}
\end{equation}
\begin{equation} \label{eq:14}
    h_i^{(l+1)}=GRU(m_i^{(l+1)},h_i^{(l)})
\end{equation}

\begin{figure}[!ht]
  \centering
  \includegraphics[width=15cm,height=11cm]{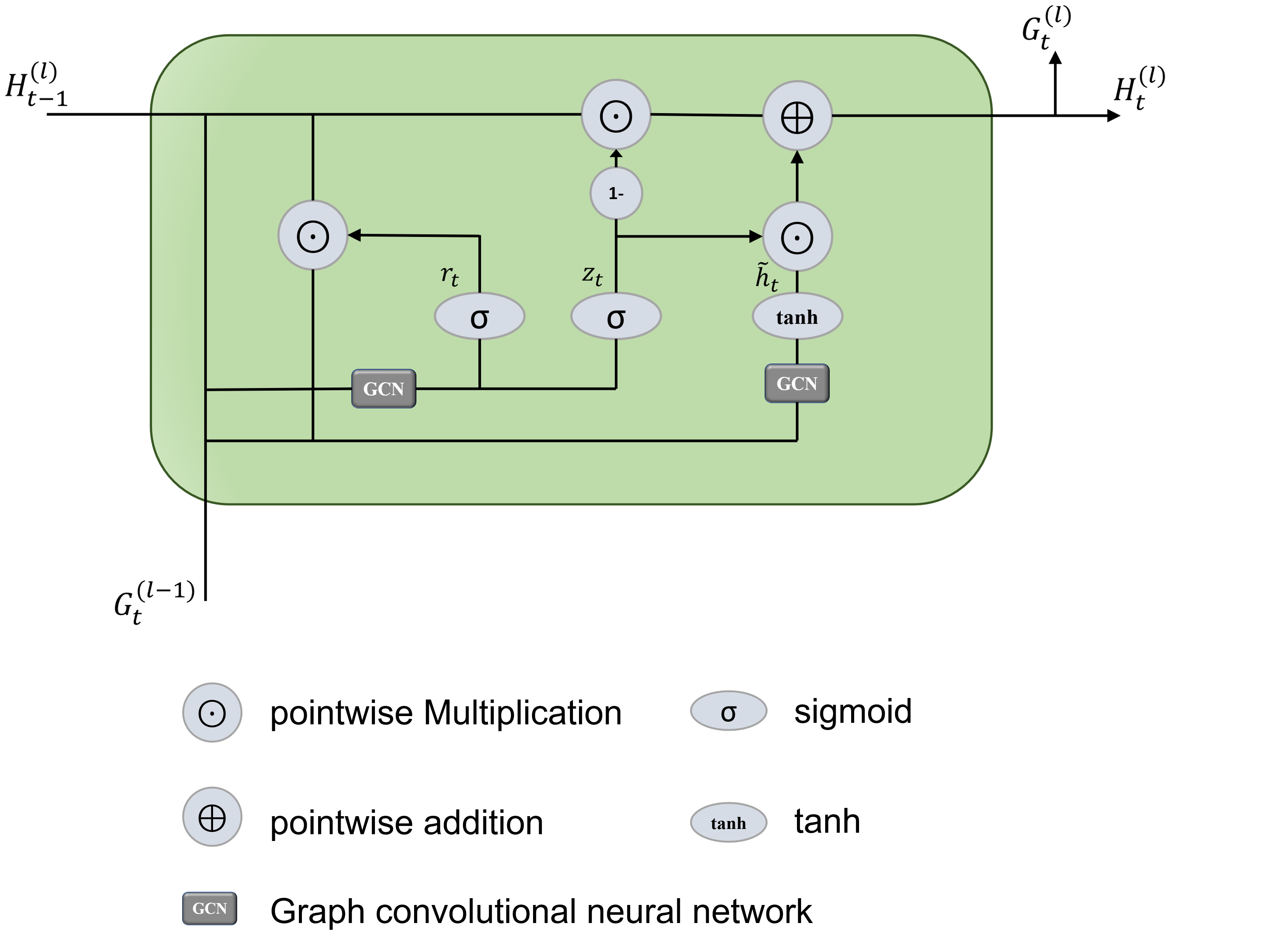}
  \caption{Gate graph convolution neural network  structure} \label{fig:GGCN}
\end{figure}

If we want to expand equations \autoref{eq:12} to \autoref{eq:14} expressed as follows:

\begin{equation}
    h_v^{(1)}=[x_v^T , 0]
\end{equation}
\begin{equation}
    a_v^{(t)}=A_{v:}^T[h_1^{(t-1)^T}, \dots ,h_{|v|}^{(t-1)^T} ]^T + b
\end{equation}
\begin{equation}
    z_v^t = \sigma (W^z a_v^{(t)} + U^z h_v^{(t-1)})
\end{equation}
\begin{equation}
    r_v^t = \sigma (W^r a_v^{(t)} + U^r h_v^{(t-1)})
\end{equation}
\begin{equation}
    \tilde{h_v^{(t)}}=\tanh{(Wa_v^{(t)}+U(r_v^t \odot h_v^{(t-1)})}
\end{equation}
\begin{equation}
    h_v^{(t)}=(1 - Z_v^t) \odot h_v^{(t-1)} + Z_v^t \odot \tilde{h_v^{(t)}}
\end{equation}
Where $U \in \mathbf{R^{NxN}}$ is the weight hidden state parameters to be learned. As a means of better understanding, \autoref{fig:GGCN} presents an overview of the structure of GGCN (Graph Gated Convolutional Network). At its core, GGCN utilizes GRU (Gated Recurrent Unit) as its basic building block. However, in GGCN, the traditional linear connections in GRU are replaced with GCN (Graph Convolutional Network) connections.

To estimate the demand at stations, a series of GGCNNs has been applied along the temporal axis to capture the spatial and temporal correlations simultaneously. This method requires fewer temporal iterative operations and avoids the error accumulation issue present in RNNs. Additionally, by adopting the sigmoid function, the suggested GGCNN with the gated units is able to capture the non-linearity, in line with many previous studies, for example, \citep{dauphin2017language,cui2019traffic}.

A travel demand forecasting model is trained with the objective of minimizing the discrepancy between actual and predicted travel demand values. As a result, the loss function is defined as the mean squared error for N time steps, as follows:

\begin{equation}
    MSE=\frac{1}{N} \Sigma_{i=1}^N (\hat{y}_i - y_i)^2
\end{equation}

\begin{figure}[!ht]
  \centering
  \includegraphics[width=15cm,height=10cm]{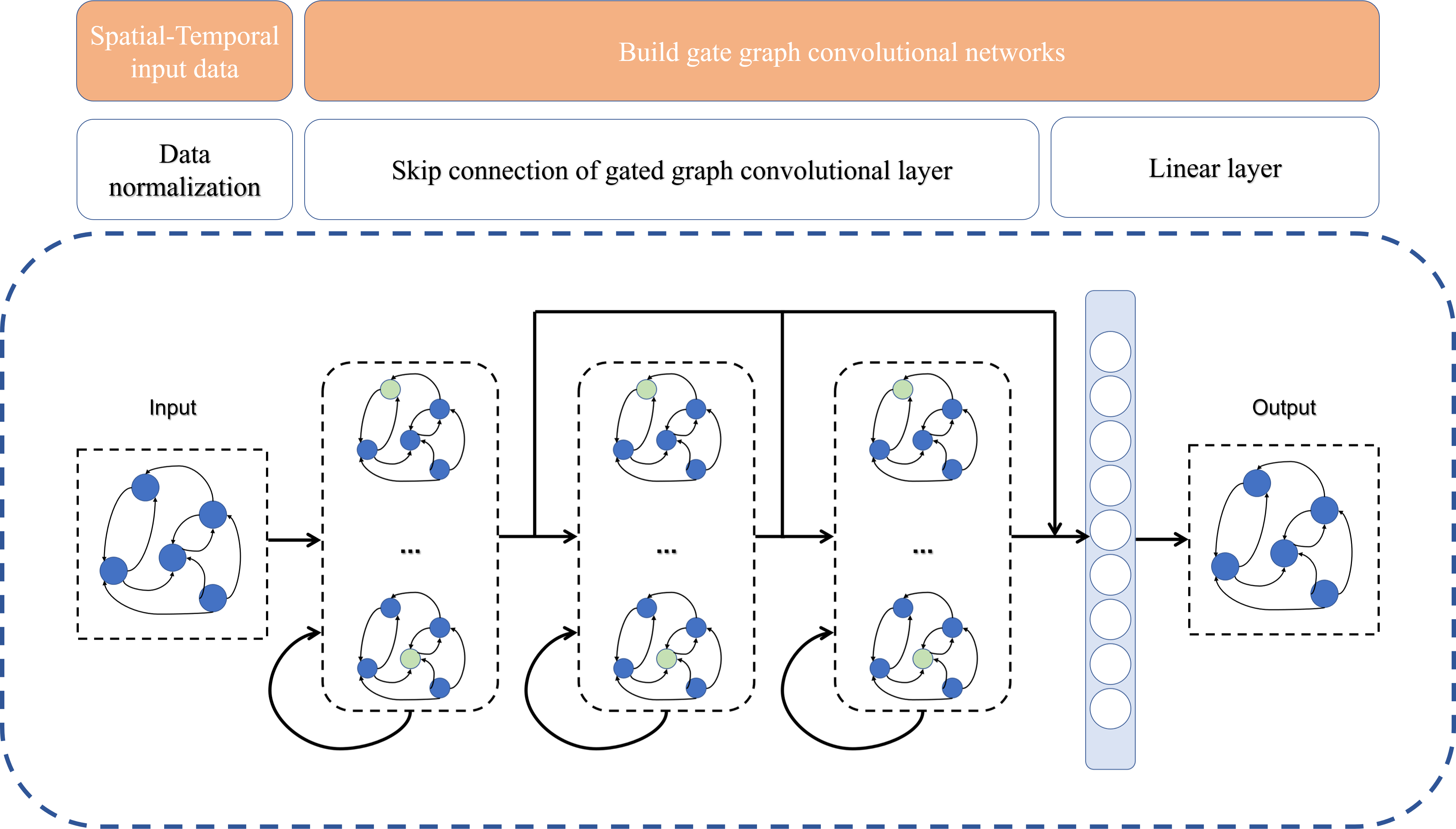}
  \caption{The framework of the proposed demand bike share prediction} \label{fig:GGCN_framework}
\end{figure}

Where $N$ number of stations $\hat{Y}= \{\hat{y_1}, \hat{y_2}, \hat{y_3}, \dots, \hat{y_N} \} $ is predicted of travel demand and $Y= \{y_1, y_2, y_3, \dots, y_N \} $ is the actual value of travel demand and $\Bar{y}$ is average actual travel demand.\\ \autoref{fig:GGCN_framework} illustrates the main prediction framework proposed in this paper. The framework comprises two crucial components: a deep GGCN-based feature representation module and a regression module. These components play a pivotal role in accurately predicting the desired outcomes.

\subsection{Locational access}
Locational access, which represents cumulative opportunities, calculates the attainable locations from a specific point. A map area displaying locational access provides valuable information on how access to various locations is dispersed, identifies areas in need of services, and highlights potential opportunities for growth and investment\citep{levinson2016accessibility}. 
The number of opportunities, such as residents or employment, within the isochrone of the query point is equivalent to the value of locational access. The query point, also known as the point of interest, could be a station gate or platform, for example. An isochrone in a transportation network is the broadest coverage provided by one or more modes from a transit node within a predetermined time window. One metric of opportunity access is the proportion of residents in an isochrone's vicinity who can walk 15 minutes to the platform of a train station \citep{lahoorpoor2020catchment}. The location access analysis is represented in  \autoref{eq:1}

\begin{equation}
\label{eq:1}
    A_{i,T}=\Sigma_{j=1}^J O_j . f(C_{ij})
\end{equation}

There is a cumulative opportunity $A_{i,T}$ between station (i) and every other location reachable in time T; At location j, $O_j$ represents the number of opportunities (population or jobs); $C_{ij}$ is walking time from station i to location j; $f(C_{ij})$ is equal to 1 if $C_{ij} \leq T$ and $0$ otherwise. 

Both continuous and discontinuous spatiotemporal networks can calculate isochrones. While isochrones for public transportation systems are discrete in both time and space dimensions, walking isochrones are in a continuous space where all points are accessible at any moment \citep{gamper2011defining}. In this study, only walking isochrones from each bike-sharing station are considered.
\subsection{Evaluation metrics}
In this study, to assess the performance of the models, several evaluation metrics were employed, including Root Mean Squared Error ($RMSE$), Mean Squared Error ($MSE$), and R-squared ($R^2$). These metrics were calculated as follows, providing a comprehensive measure of the model's accuracy and predictive capability:

\begin{equation}
    RMSE=\sqrt{\frac{1}{N} \Sigma_{i=1}^N (\hat{y}_i - y_i)^2}
\end{equation}

\begin{equation}
    R^2=1-\frac{\Sigma_{i=1}^N(y_i-\hat{y_i})^2}{\Sigma_{i=1}^{N}(y_i-\Bar{y})^2}
\end{equation}
where $N$ number of stations $\hat{Y}= \{\hat{y_1}, \hat{y_2}, \hat{y_3}, \dots, \hat{y_N} \} $ is predicted of travel demand and $Y= \{y_1, y_2, y_3, \dots, y_N \} $ is the actual value of travel demand and $\Bar{y}$ is average actual travel demand.

\subsection{Base models} \label{sec:Base models}
In order to evaluate the effectiveness of the main model, a comparative analysis was conducted against other models discussed in the existing literature. This study presents a comprehensive comparison between the main model and benchmark models mentioned in the literature. The following section provides a detailed explanation of the underlying mechanisms and functionalities of the base models considered in this study. By examining these models, we gain a better understanding of their strengths and limitations, which allows for a comprehensive evaluation of the main model's performance.

\textbf{Ordinary Least Square Regression (OLS):}
The Gauss-Markov theorem states that the Ordinary Least Squares (OLS) estimator is optimal among the class of linear unbiased estimators when the errors exhibit homoscedasticity and serial uncorrelation. This holds true for level-one fixed effects when the regressors are exogenous and demonstrate perfect collinearity, subject to the rank condition. Additionally, the theorem guarantees that the variance estimate of the residuals remains constant when the regressors possess finite fourth moments. Furthermore, the OLS estimator remains consistent for level-one fixed effects under various other conditions.

\textbf{Auto Integrate Moving Average(ARIMA):}
The ARIMA model is a data-oriented strategy that utilizes the inherent structure of the data. However, it faces limitations when applied to sizable nonlinear datasets. Employing ARIMA to model linear patterns can inadvertently distort nonlinear patterns. To address this, Auto ARIMA automatically generates optimal parameter values, alleviating the need for manual selection.

In the Auto ARIMA framework, the parameter $p$ represents the sub-order of the Auto Regression (AR) model. To generate accurate predictions, the model requires a certain number of historical values.

The parameter $d$ signifies the level of differencing applied to eliminate non-stationary components within the data.

The parameter $q$ relates to the sub-order of the Moving Average (MA) model, which influences the maximum number of past errors considered by the ARIMA model during the forecasting process. It plays a crucial role in minimizing forecasting errors and enhancing the overall accuracy of the model.

\textbf{Multi-layer Perceptron(MLP): }In a multi-layer perceptron artificial neural network, the design of the network architecture, such as the number of nouns or layers, plays a crucial role as it directly impacts the neural network's ability to solve a given problem. Therefore, careful consideration must be given to the design phase to ensure optimal performance. One crucial stage in employing machine learning methods, including the perceptron artificial neural network, is the training phase. During this stage, the network undergoes a process of adjusting its parameters, particularly the network weights. These adjustable parameters are optimized through mathematical optimization techniques, allowing the neural network to solve the problem at hand effectively.

\textbf{Convolution neural network (CNN):} A 1D-CNN model, consisting of a convolutional hidden layer that operates on a 1D sequence, is commonly referred to as a one-dimensional CNN. The output of the convolutional layer is subsequently condensed by a pooling layer to extract the most crucial components. In certain cases, such as when dealing with lengthy input sequences, a second convolutional layer may follow the pooling layer. Following the convolutional and pooling layers, a dense, fully connected layer is employed to interpret the features extracted by the convolutional section of the model.

\textbf{XGboost: } A widely used and effective machine learning technique is tree boosting. In this study, we employed the scalable end-to-end tree boosting technique known as XGBoost was utilized, which is favored by data scientists for delivering state-of-the-art results across various machine learning problems \citep{chen2016xgboost}.
 
Gradient boosting, a class of ensemble machine learning techniques, is employed for classification or regression predictive modeling tasks. Ensembles are constructed from decision tree models. Trees are sequentially added to the ensemble to rectify prediction errors made by earlier models. This iterative process of adding and fitting trees characterizes the boosting approach in ensemble machine learning.

The fitting of models can be accomplished using the gradient descent optimization approach along with any differentiable loss function. The term "gradient boosting" is derived from the fact that the loss gradient is minimized during the model fitting process, similar to how it occurs in a neural network.

\textbf{Long Short-Term Memory (LSTM):} 
The LSTM architecture was developed specifically to address the challenge of handling long-term dependencies. It was inspired by studying the error flow within functioning RNNs. Through this examination, it was observed that existing architectures struggled with long-time lags due to the backpropagated error either exponentially decaying or amplifying.

To overcome this limitation, the LSTM architecture was introduced. It enables RNNs to effectively capture and retain information over extended sequences, preventing the vanishing or exploding gradient problem. This makes LSTM particularly suitable for tasks where understanding and modeling long-term dependencies are crucial \citep{graves2005bidirectional}.

\textbf{Support Vector Regression (SVR): }SVR utilizes a sparse kernel, a classification technique that involves a hyperplane defined by a small set of support vectors. This concept of using support vectors as the basis for classification is the fundamental principle of SVM (Support Vector Machines). In SVR, the sparse kernel enables efficient regression by identifying the essential support vectors contributing significantly to the regression model. By utilizing only a subset of support vectors, SVR can effectively model complex relationships while maintaining computational efficiency \citep{zhang2020support}.

\textbf{Graph Convolution Neural Network (GCN): }In this paper, the GCN neural network is used to compare with the main model; the structure of the model GCN is given in the method section.

\textbf{GCN-LSTM: }
The nature of numerous real-world phenomena involves spatial and temporal dynamics. The approach undertaken involves encoding and decoding the structure by embedding a Graph Convolutional Network (GCN) unit within the Long Short-Term Memory (LSTM) unit. This integration facilitates the simultaneous capture of spatial and temporal features \citep{lin2018predicting}.

\textbf{Attention-Based Spatial-Temporal Graph Convolution model (ASTGCN): }
Attention is a rapidly developing mechanism that focuses on input data to prioritize crucial information for prediction effectively. The application of attention mechanisms in time series data involves focusing on specific time points or series of time. In the context of graph convolution, the attention mechanism assigns weights to important neighboring points, significantly influencing predictions during data processing. Investigating and utilizing the attention mechanism in Graph Convolutional Networks (GCN) constitutes one of the recent methods employed in investigations, and we assume that the graph is dynamic \citep{kong2020stgat}.

\textbf{Multi Attention Based Graph Convolution model(MGAT): }
The multi-graph attention network offers several advantages, including the ability to handle heterogeneous relationships. The model can capture diverse aspects of a database, thereby enhancing overall performance. One crucial reason to employ this architecture is its capacity to preserve critical information by separately capturing relationships. The network architecture consists of multiple independent components within the Attention Graph Network, depicting heterogeneous relationships between stations. Specifically, it employs an encoder-decoder structure, where the output of the encoder is forwarded into a fully connected network to predict travel demand \citep{kong2020stgat}.

\section{Case study}
\label{sec:case}

The city of Chicago is located in the state of Illinois in the United States. Divvy bike share is one of the features of Chicago's transportation system. Launched in 2013, Divvy provides individuals with a convenient and environmentally friendly means of transportation throughout the city. By providing a network of bicycle stations throughout the city, users are able to easily rent bicycles and return them to any station within the system. Chicago's Divvy bike share program has become an integral part of the city's transportation infrastructure, promoting sustainability and reducing traffic congestion. Since its inception, the Divvy bike share program in Chicago has experienced growth and popularity \citep{faghih2015analysing}.

To investigate the travel demand for Divvy bike-share, this study collected historical data from various sources. The ride-sourcing trajectory data for Chicago was obtained from the publicly accessible Divvy Bikes dataset\footnote{Access to the data can be found at the following link: \href{https://divvybikes.com/}{https://divvybikes.com/}}. Additionally, historical weather data was extracted from the openweather database\footnote{Access to the data can be found at the following link:\href{https://openweathermap.org/}{https://openweathermap.org/}}, while demographic information was sourced from the Chicago Data Portal\footnote{Access to the data can be found at the following link:\href{https://www.census.gov/}{https://www.census.gov/}}.

This study utilized three distinct types of datasets as features for the analysis: bike-sharing transactions, weather conditions, and Census data. Each database contributed unique information that helped understand travel demand patterns. In the subsequent sections, we will explore the specific features and characteristics of each dataset and eventually describe the preprocessing of the database.

\subsection{Bike-sharing transaction}
The study dataset comprises more than $3$ million bike-sharing transactions that occurred in Chicago in 2019. These transactions were acquired from Divvy, the bike-sharing service in the city. Each transaction record provides valuable information such as the duration of the trip, check-in and check-out times of the bike, names of the starting and ending stations, latitude and longitude coordinates of the stations, user ID, user type (Customer or Subscriber), and other relevant details. The dataset includes information from multiple stations, and for the purpose of this study, these stations are considered for analysis. To capture the relationship between the stations, an adjacency matrix is constructed using the known travel demands between them. This matrix enables the study to examine the connectivity and interaction patterns among the different stations in the bike-sharing network.

\begin{figure}[!ht]
  \centering
  \includegraphics[width=15cm,height=10cm]{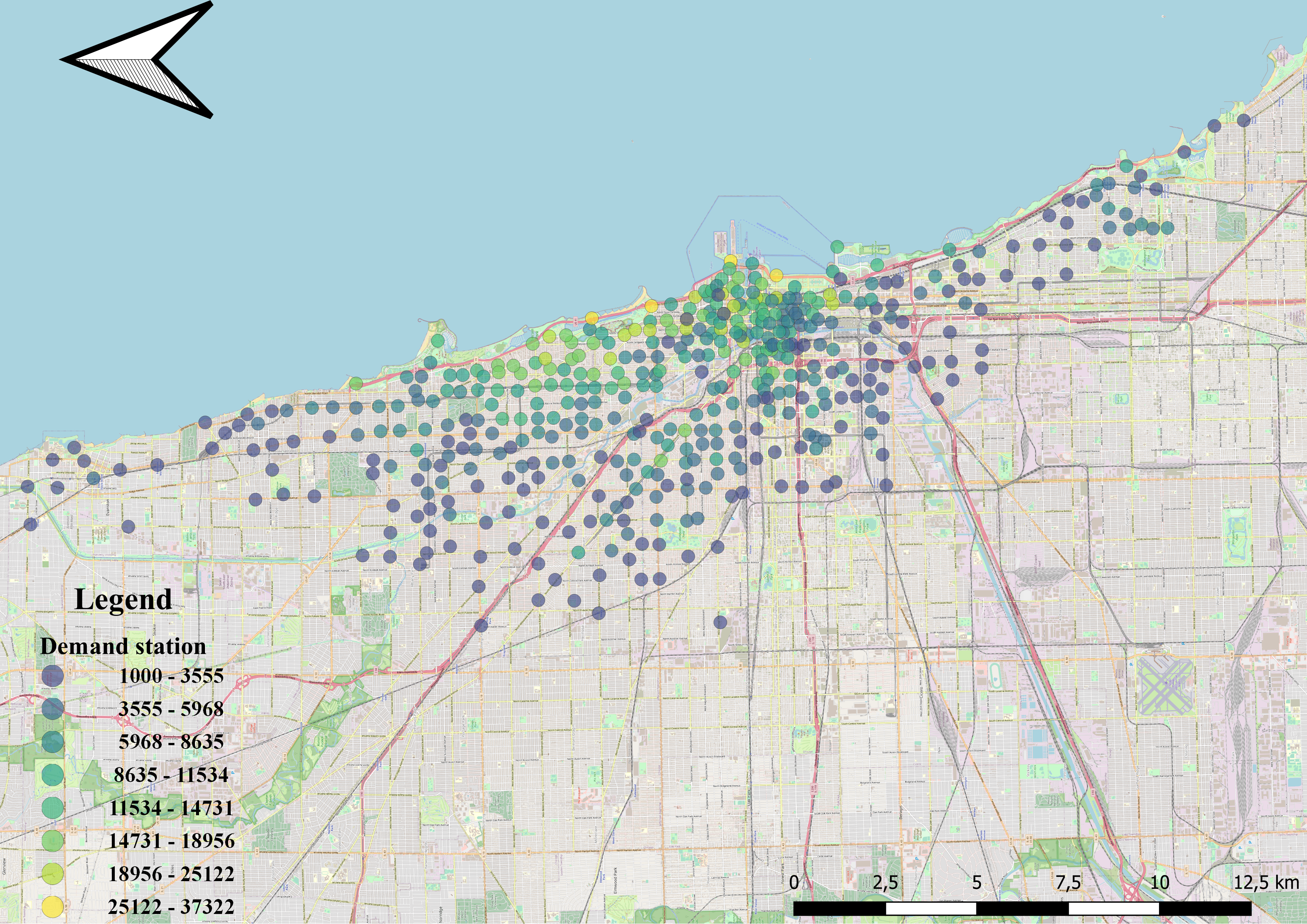}
  \caption{Stations demand in the one year} \label{fig:station_demand}
\end{figure}

The Divvy Bikes program in Chicago comprises 612 stations located in the urban area. \autoref{fig:station_demand} illustrates the coordinates of the stations that experience more than 1000 travel demands per year. In this investigation, we eliminated five stations for which no information was available, and we also filtered out stations with travel demands exceeding 1000 per year. In other words, we selected stations where there are at least two trips per day. Consequently, 500 stations were retained to predict travel demand.

\begin{figure}
     \centering
     \begin{subfigure}[!ht]{0.3\textwidth}
         \centering
         \includegraphics[width=0.9\linewidth, height=6cm]{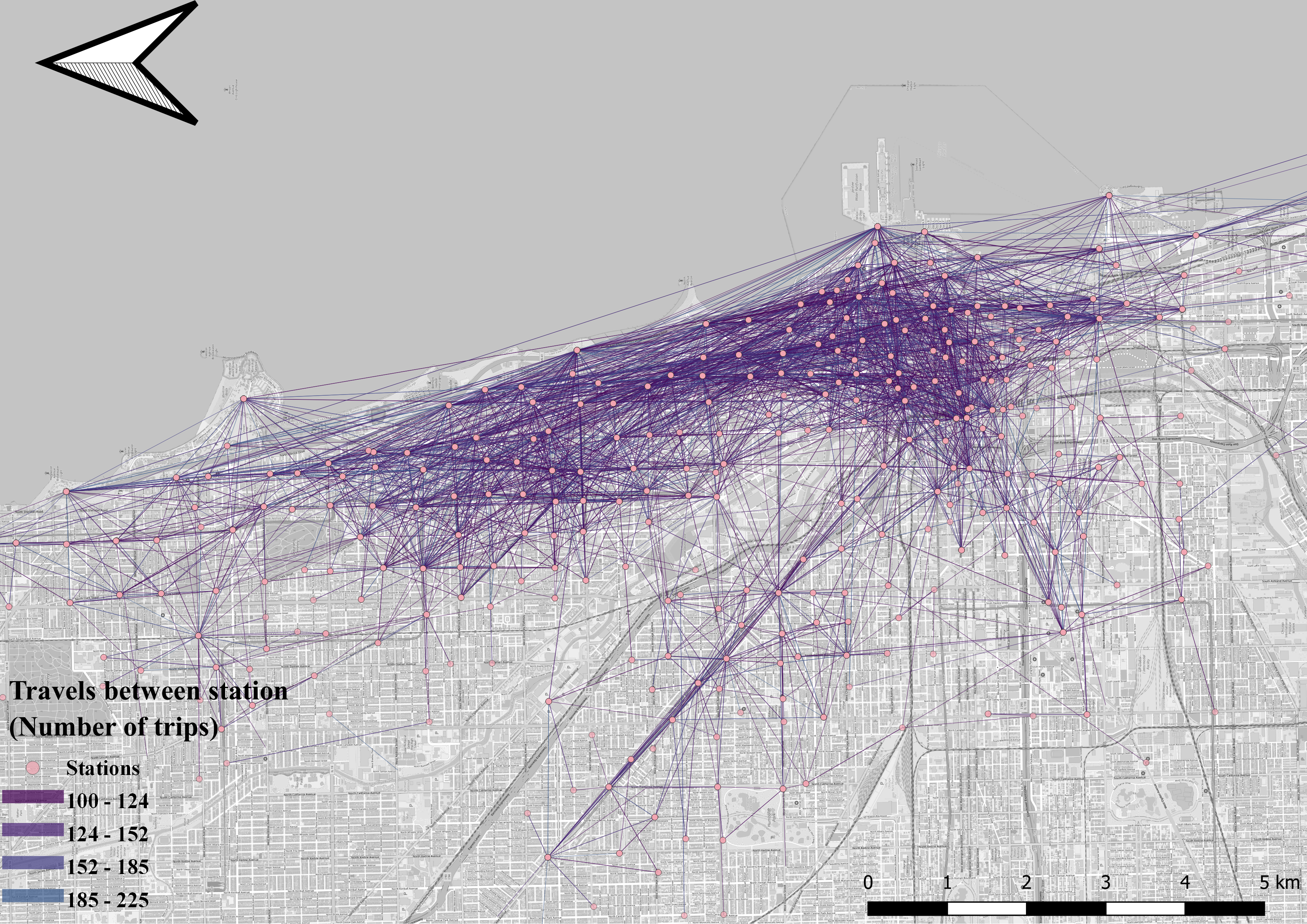}
         \caption{}
         \label{fig:transaction_1}
     \end{subfigure}
      \hspace*{3mm}
     \begin{subfigure}[!ht]{0.3\textwidth}
         \centering
         \includegraphics[width=0.9\linewidth, height=6cm]{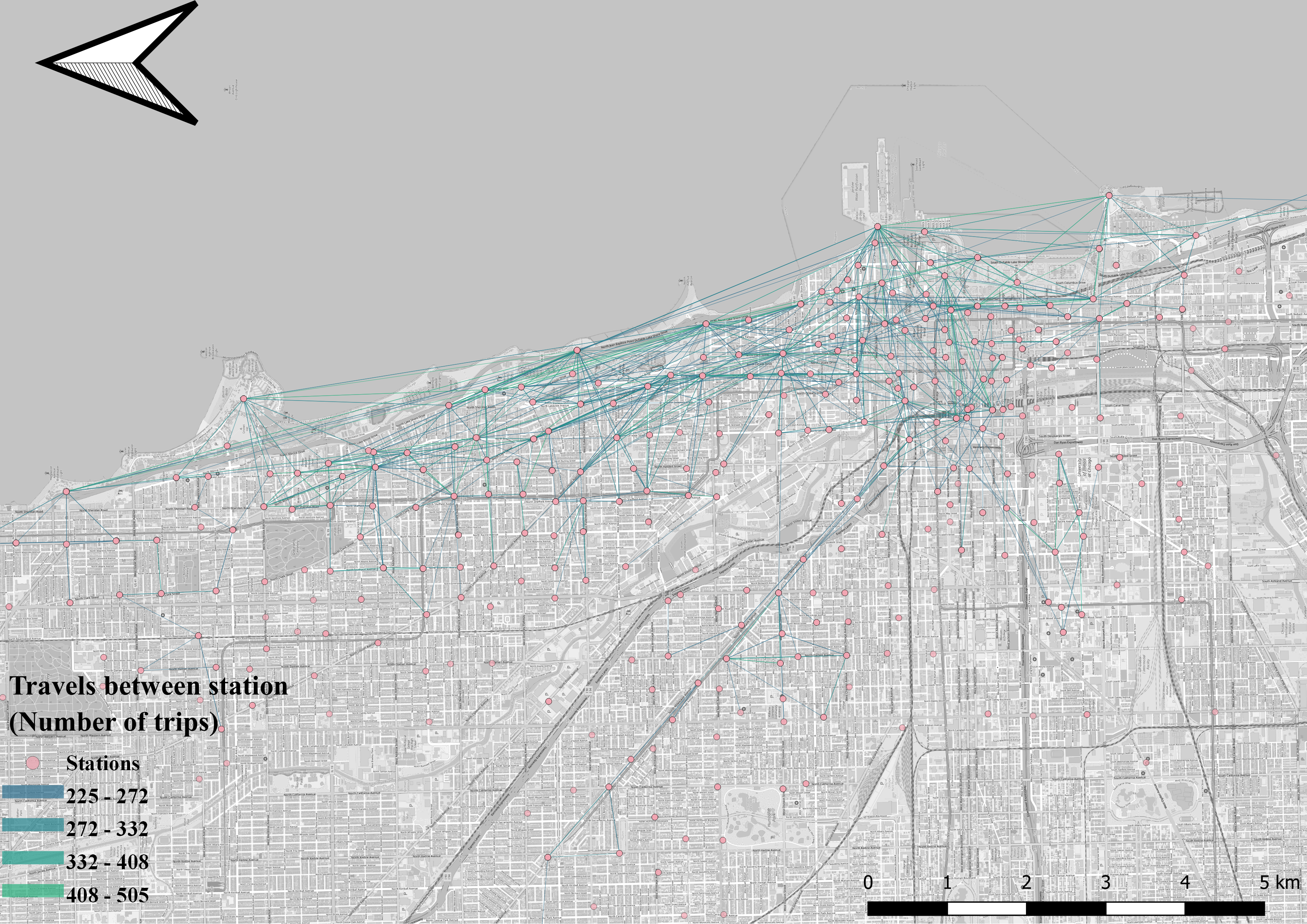}
         \caption{}
         \label{fig:transaction_2}
     \end{subfigure}
     \hspace*{3mm}
     \begin{subfigure}[!ht]{0.3\textwidth}
         \centering
         \includegraphics[width=0.9\linewidth, height=6cm]{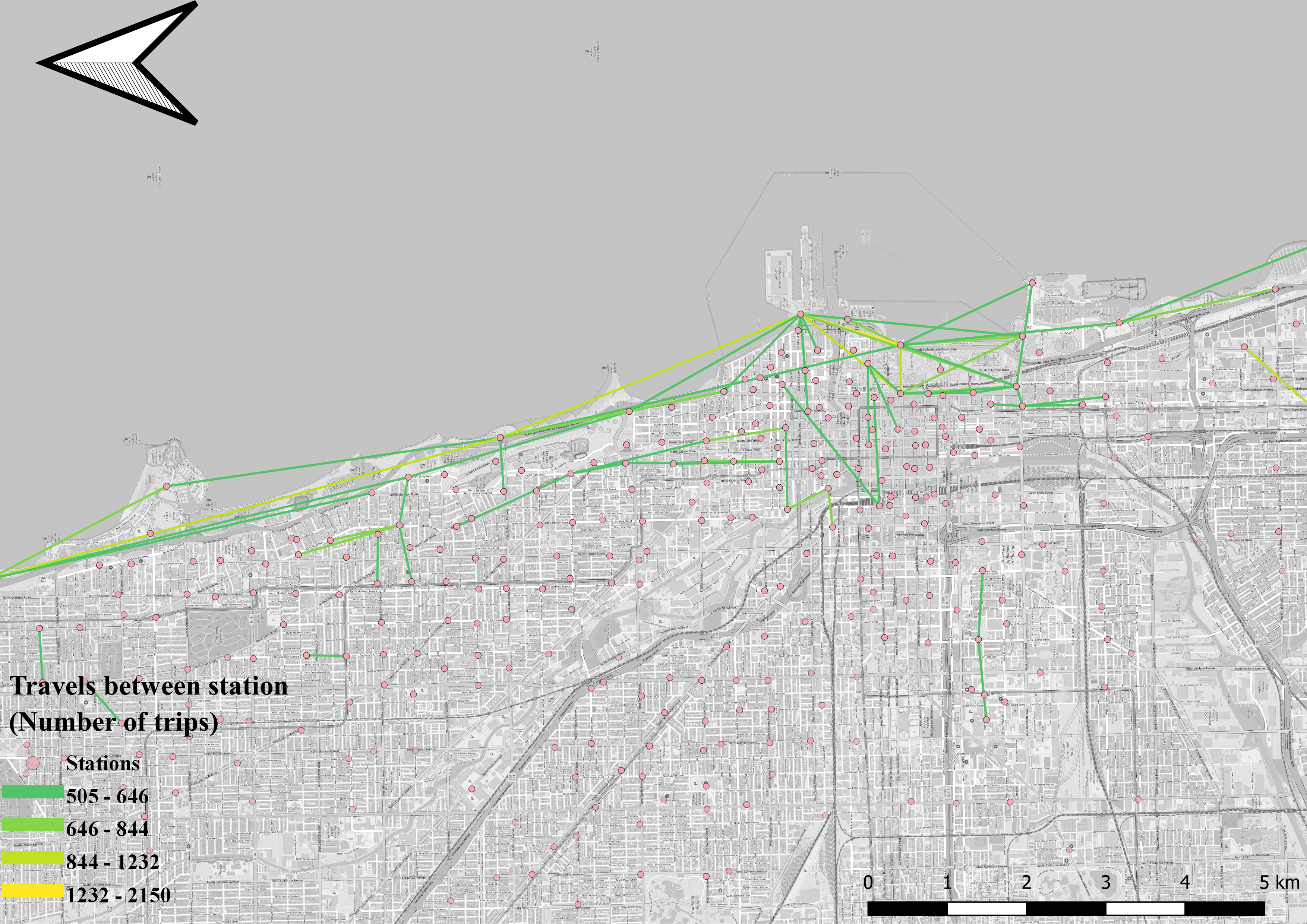}
         \caption{}
         \label{fig:transaction_3}
     \end{subfigure}
     \caption{Transactions bike-sharing between selection stations in a year (a) Transactions bike-sharing between 100 - 225 (b) Transactions bike-sharing between 225 - 505 (c) Transactions bike-sharing between 505 - 2150}
 \end{figure}

The traffic flow between selected stations is depicted in three figures: \autoref{fig:transaction_1} illustrates the number of transactions between 100-225, \autoref{fig:transaction_2} showcases the number of transactions between 225-505, and \autoref{fig:transaction_3} demonstrates the number of transactions between 505-2150. These figures provide visual representations of the traffic volume between specific station pairs, offering valuable insights into the patterns and intensity of movement within the transportation network. The most transactions have been in stations Lincoln Park Conservatory, Michigan Ave, and Marine Dr \& Ainslie St.

\subsection{Weather condition historical data}
One of the aims of this study is to investigate the impact of weather conditions on travel demand at each station. To accomplish this, historical weather data with a one-hour time step was obtained from the website OpenWeatherMap. Each weather condition record in the dataset includes information such as the date and time (year/month/day hour: minutes: seconds), temperature, wind speed (km/h), humidity, precipitation (mm), pressure, total snowfall (cm), cloud cover, and more. These comprehensive weather variables provide essential insights into the prevailing atmospheric conditions during the study period. By analyzing the relationship between these weather factors and the corresponding travel demand, the study aims to uncover how weather influences the usage patterns of the bike-sharing system.

\begin{figure}[!ht]
  \centering
  \includegraphics[width=16cm,height=12cm]{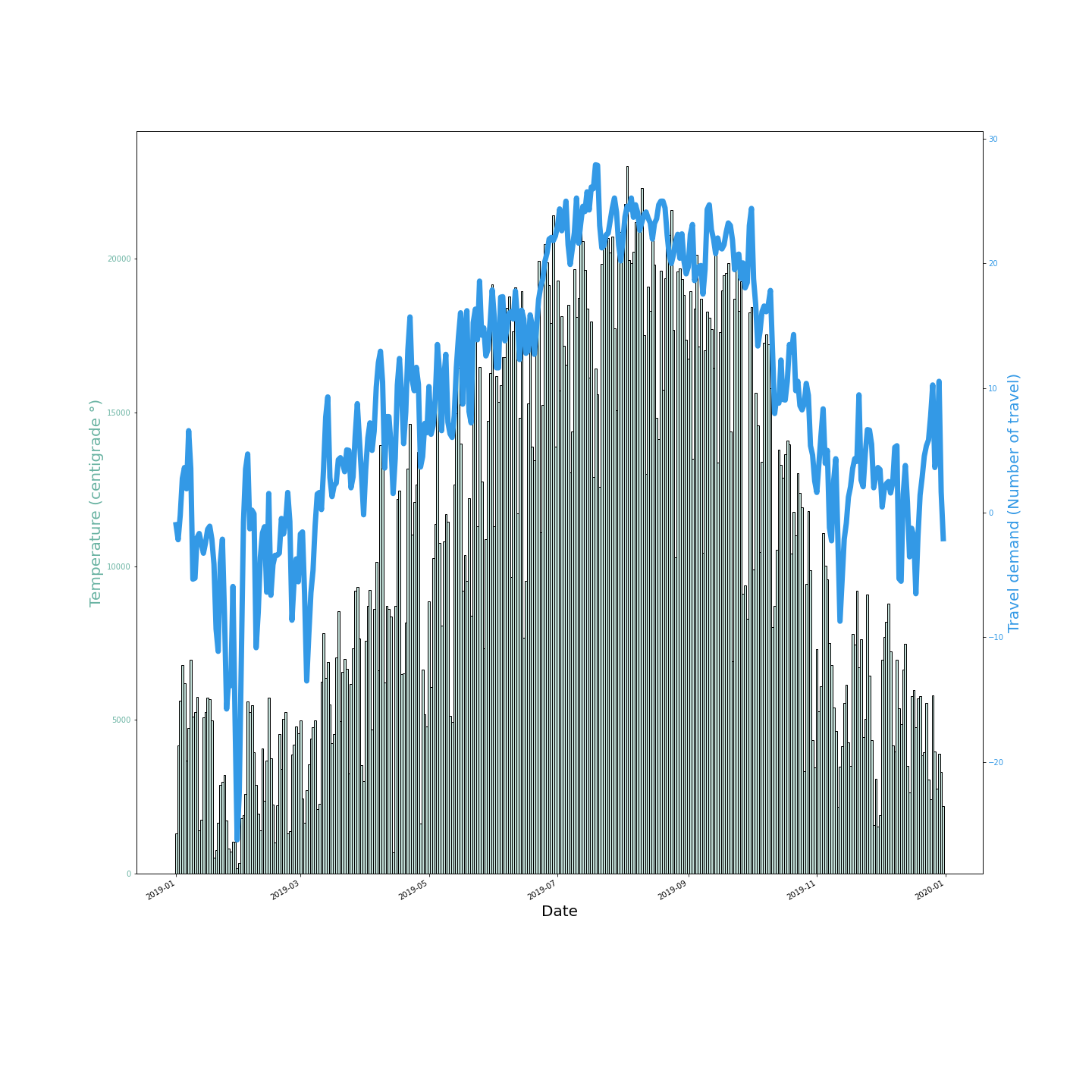}
  \caption{Impact of the temperature on demand} \label{fig:impact temperature on demand}
\end{figure}

\autoref{fig:impact temperature on demand} illustrates the influence of temperature on travel demand. It is evident that there is a direct relationship between the amount of travel demand and temperature. As the temperature rises, the utilization of the travel demand also increases. This finding highlights the sensitivity of travel demand to temperature variations.

\subsection{Census data}
Census data for Chicago's population is currently available from 2010 when the city's population was approximately 3 million. Despite the possibility of minor changes in the distribution and density of population and jobs since then, due to the lack of updates in recent years, the 2010 census data is used in this study, as well as the latest employment data for Chicago available as of 2019.

In order to determine access, the number of people and businesses in each location within the isochrone must be summed. The Chicago Census data consists of census blocks, which are the smallest geographical areas designed more as geographical building blocks than as areas in which statistics are released themselves, and include both the census block and the total population.

\begin{figure}[!ht]
  \centering
  \includegraphics[width=16cm,height=11cm]{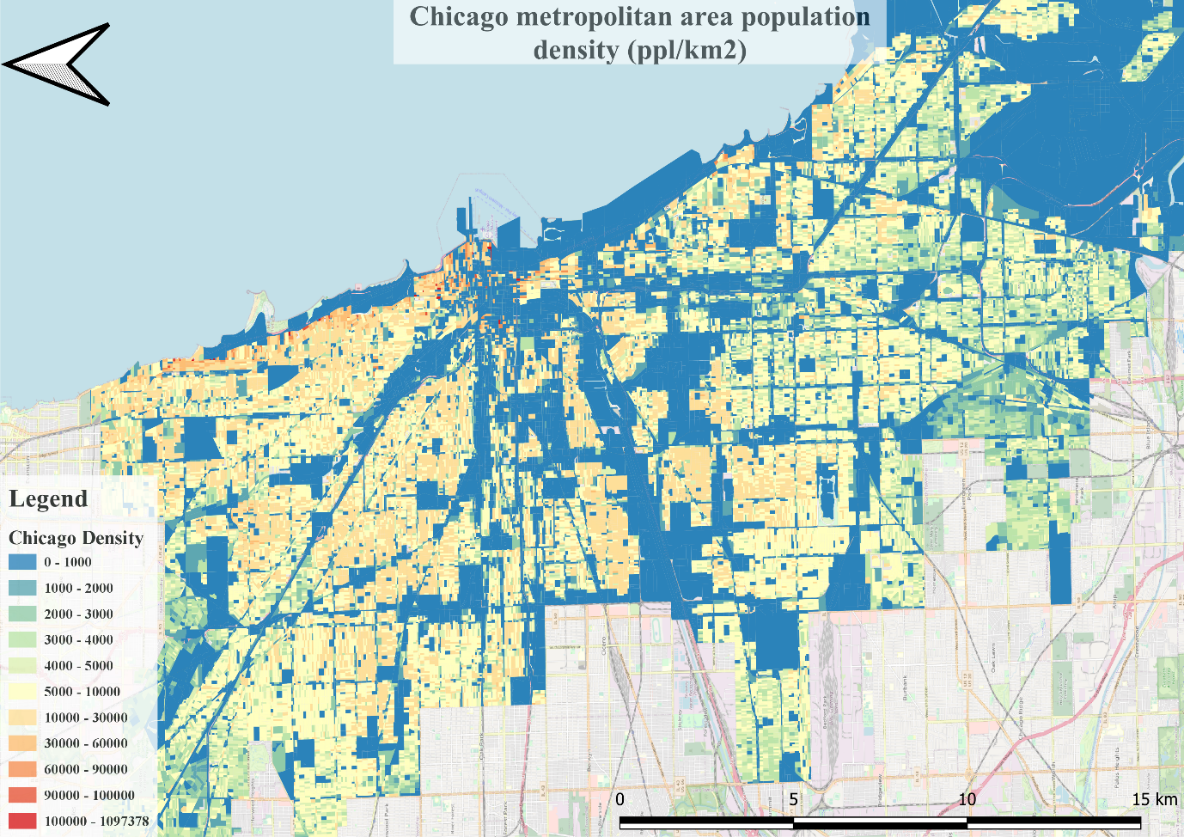}
  \caption{Chicago population density} \label{fig:Density}
\end{figure}

Areas with higher population density have a greater impact on travel demand compared to areas with lower population density. The density of population refers to the amount of population within a given land area in each zone. The reference to \autoref{fig:Density} illustrates the population density in Chicago, showcasing the distribution and concentration of people across different areas.

Regarding the availability of the latest employment data accessed in 2019, census population data from 2010, and the impact of COVID-19 on travel patterns in 2020, we aimed to compare our framework with other studies from the literature. Consequently, we selected transaction data from 2019.

\subsection{Preprocessing data}
The datasets were processed using the following procedure to prepare them for demand modeling. First, since a significant proportion of trips according to trip ID and user ID were recorded, the number of pickup bikes was based on each hour on each date at each geography coordinate station by applying the Python ''Pandas and Numpy'' package grouping. 

Second, we summarized the trajectory trip data from a disaggregate level (i.e., individual trip records) into an aggregate level (i.e., travel demand at each station). We calculated the following new attributes: the total number of in-trips and out-trips in each hour, and access at each station. In the next step, we merged the historical weather dataset with a one-hour time step with the main dataset. Additionally, We applied the Min-Max scaling method to the datasets. Finally, we used this combined dataset to develop spatial-temporal of bike-sharing demand in Chicago. For data splitting, we separate 30 percent of the entire dataset as the test data, which remains constant for comparing all the base models with the main model. The remaining 70 percent of the data is used for training and evaluation purposes.

\autoref{tab:variable model} presents statistical features such as the minimum (min), maximum (max), standard deviation (std), and mean for a set of introduced features. The table provides a summary of these statistical measures for a specific set of features or variables. With the increased number of features as input for the model, there is a notable rise in computational demand and the risk of overfitting \citep{ying2019overview}. To address these challenges, it becomes imperative to explore statistical features. To alleviate these issues, it is essential to eliminate features that demonstrate redundancy, share similar traits, or exhibit high correlation with other features. Moreover, during the data processing phase, it is crucial to identify and remove features characterized by outlier distributions.

Following a meticulous analysis of the processed data and an exploration of statistical features, we have identified that the specific features detailed in \autoref{tab:variable model} are well-suited for integration into our framework. In subsequent research endeavors, researchers must acquire a comprehensive understanding of our database while implementing and advancing our framework.

\begin{table}[!ht]
\setlength{\tabcolsep}{10pt} 
\renewcommand{\arraystretch}{1.5} 
    \centering
    \caption{The descriptive statistics of the variables used in the model}
    \begin{tabular}{lllll}
    \toprule
      Variable  & Mean  &  std  & Min & Max\\
    \midrule
 
         Humidity  & 73.601 & 12.638 & 29 & 100 \\         
         Precipitation & 0.245 & 0.751 & 0 & 16.300 \\     
         Pressure & 1016.902 & 7.544 & 991 & 1048 \\     
         Temperature & 9.516 & 10.627 & -28 & 31 \\     
         Wind speed & 16.244 & 6.898 & 1 & 45 \\     
         access to population & 17134.51 & 8057.615 & 0 & 37747 \\     
         access to employment & 884.134 & 945.173 & 0 & 4592 \\     
         Last time-step in-trip & 1.76 & 4.03 & 0 & 153 \\     
         Last time-step out-trip & 1.77 & 4.01 & 0 & 149 \\     

    \bottomrule
    \end{tabular}
    \label{tab:variable model}
\end{table}

\section{Result}
\label{sec:result}

To evaluate the performance of the proposed model, a comparison is made with the base models presented in \autoref{sec:Base models}. This section provides insights into the architecture, sample size, features utilized, and accuracy achieved by each model. By comparing the proposed model with these baseline models, we aim to showcase the superior performance and effectiveness of our approach. 

\textbf{Ordinary Least Square Regression (OLS):}
We used approximately 1,000,000 samples, each comprising an input-target pair. The input data is structured as a 2-dimensional matrix with dimensions (samples, features). Each input part of the input-target pairs consists of eight features, including precipitation, pressure, temperature, wind speed, access to population and employment areas, and the number of in-trips and out-trips in the last time step.

After training the model on this dataset, we evaluated its performance using various metrics. The results indicate an R-square value of 0.5, and the mean squared error (MSE) is calculated to be 0.49: furthermore, the root mean squared error (RMSE) is measured at 0.7.

\textbf{Auto Integrate Moving Average(ARIMA):}
We generated approximately 1,000,000 samples, consisting of input-target pairs, where the input shape is a 2-dimensional matrix (samples, features). Each input part of the input-target pairs includes eight features: precipitation, pressure, temperature, wind speed, access to population and employment areas, as well as the number of in-trips and out-trips in the last time step.

After training the model on this dataset, we evaluated its performance using various metrics. The results indicate an R-square value of 0.1, and the mean squared error (MSE) is calculated to be 5.6; furthermore, the root mean squared error (RMSE) is measured at 2.37.

\textbf{Multi-layer perceptron(MLP): }
We generated approximately 1,000,000 samples, comprising input-target pairs with a 2-dimensional matrix shape (samples, features). Each input part of the input-target pairs consisted of eight features, including precipitation, pressure, temperature, wind speed, access to population and employment areas, as well as the number of in-trips and out-trips in the last time step.

The MLP model used in this study consisted of three layers with 32, 32, and 64 neurons, respectively, and employed the ReLU activation function. The loss function utilized was the mean squared error, while the optimizer employed was stochastic gradient descent (SGD), the learning rate was 0.001, and the batch size was 64. 

After training the model on this dataset, we evaluated its performance using various metrics. The results indicate an R-square value of 0.66, and the mean squared error (MSE) is calculated to be 0.11; furthermore, the root mean squared error (RMSE) is measured at 0.32.

\textbf{Convolution neural network (CNN):}
The feature maps are condensed into a single one-dimensional vector through the use of a flatten layer positioned between the dense and convolutional layers. In this research, the input travel demand for each station is recorded at a time step of 1 hour, resulting in a 3-dimensional matrix with the dimensions (sample size, time step, number of stations). To filter the data, we selected 153 stations that had a demand exceeding 5000 throughout the year. The model's input feature comprises a sequence of travel demand with a length of 24.

The model implementation structure comprises two layers of 1-dimensional convolution with a kernel size of 2, followed by max pooling. The flatten layer is then employed to transform the output into a one-dimensional vector. Finally, a fully connected layer is added to complete the model architecture. The investigation utilized the Adam optimizer with a learning rate of 0.001, a batch size of 32, and employed the mean square loss function. The training process comprised 100 epochs, and ultimately, we employed the ReLU activation function.

After training the model on this dataset, we evaluated its performance using various metrics. The results indicate an R-square value of 0.11, and the mean squared error (MSE) is calculated to be 2.30; furthermore, the root mean squared error (RMSE) is measured at 1.24.

\textbf{Long short-term memory (LSTM): } 
The input shape of the model is a 3-dimensional matrix, which includes the dimensions of sample size, time step, and number of stations. In this study, we filtered out 153 stations that exhibited a demand of over 5000 throughout the year.

The model implementation structure consists of two layers of LSTM, with the number of units set at 256 and 128, respectively. Dropout layers with dropout rates of 15\% and 25\% are incorporated to mitigate overfitting, and the ReLU activation function was employed. The loss function utilized was the mean squared error, and the optimizer used was Adam with a learning rate set to 0.001, the batch size was 64 and the training process involved 100 epochs. Finally, a fully connected layer is added to the architecture to process and interpret the extracted features. The input feature of the model is a sequence of travel demands with a length of 24.

After training the model on this dataset, we evaluated its performance using various metrics. The results indicate an R-square value of 0.18, and the mean squared error (MSE) is calculated to be 2.06; furthermore, the root mean squared error (RMSE) is measured at 1.43.

\textbf{Support Vector Regression (SVR): }
We generated approximately 150,000 samples, each consisting of an input-target pair. The input data is represented as a 2-dimensional matrix with dimensions (samples, features). The eight features included in each input part of the input-target pairs are precipitation, pressure, temperature, wind speed, access to population, business, and the last time step of in-bound and out-bound values.

For the training of the model, we selected a value of 0.2 for epsilon. This value determines the size of the "error tube," which represents an area where errors are not penalized. By setting epsilon, we can effectively use data points located near the boundaries of the error tube as support vectors in the training process. Additionally, the radial basis function kernel, also known as the RBF kernel, is employed in this algorithm.

After training the model on this dataset, we evaluated its performance using various metrics. The results indicate an R-square value of 0.32, and the mean squared error (MSE) is calculated to be 0.68; furthermore, the root mean squared error (RMSE) is measured at 0.82.

\textbf{Graph Convolution Neural Network (GCN): }
In this study, the travel demand for each station was inputted with a time step of 1 (in hours) to determine the structure of the 3-dimensional matrix (Batch, Nodes, Features). We selected 500 stations with a demand of over 1000 within a year. The model implementation consists of two convolutional layers, followed by a fully connected layer. The training process involved 100 epochs with a batch size of 64, and the ReLU activation function was employed. The loss function utilized was the mean squared error, and the optimizer used was Adam, with a learning rate set to 0.001. Additionally, a min-max scaler was utilized for the matrix features. Over the course of one year, we generated an adjacency matrix for each hour, resulting in a total of 8,768 adjacency matrices, covering the entire span of 2019.

After training the model on this dataset, we evaluated its performance using various metrics. The results indicate an R-square value of 0.44, and the mean squared error (MSE) is calculated to be 0.85; furthermore, the root mean squared error (RMSE) is measured at 0.61.

\textbf{XGboost: }
We generated approximately 1,500,000 samples (input-target pairs) from a bike-sharing transaction database. The input data is represented as a 2-dimensional matrix (Batch, Features). The eight features included in each input part of the input-target pairs are precipitation, pressure, temperature, wind speed, access to population, access to employment, as well as the last time step's inbound and outbound values.

To prevent overfitting, we experimented with different values of hyperparameters. Ultimately, we settled on an ensemble size of 700 trees, with a maximum depth of 4 for each tree, and a learning rate of 0.015. These selections were made to ensure optimal performance and generalization of the model.

After training the model on this dataset, we evaluated its performance using various metrics. The results indicate an R-square value of 0.65, and the mean squared error (MSE) is calculated to be 0.35; furthermore, the root mean squared error (RMSE) is measured at 0.59.

\textbf{GCN-LSTM: }The aim of this proposed model is to capture spatial and temporal characteristics simultaneously. In this model, we filter 500 stations that exceed a demand of 1000 in a year. The input features matrix is three-dimensional (Batch, Nodes, Features). The features include precipitation, pressure, temperature, wind speed, accessibility to population, accessibility to employment, and the last time steps in-trips and out-trips. We assume the graph is dynamic, and the network architecture considers one layer of graph convolution and two layers of LSTM, with a hidden cell count of 64. Eventually, the model is connected to a fully connected layer for predicting travel demand. The training involved 100 epochs with a batch size of 64, and the ReLU activation function was employed. The loss function utilized was the mean squared error, and the optimizer used was Adam with a learning rate set to 0.001. and normalization by min-max. The total input adjacency matrix is 8,768. The composition of the input matrix, illustrating the characteristics of individual nodes, is three-dimensional, defined by the dimensions (Batch, Nodes, Features). After training, the mean squared error (MSE) is 0.42, the root mean squared error (RMSE) is 0.65, and the R-squared value is 0.32.

\textbf{Attention-Based Spatial-Temporal Graph Convolution model (ASTGCN): }
The main idea of the proposed model revolves around the attention mechanism, which extracts neighboring nodes crucial for forecasting travel demand. The core of the model is the Graph Convolutional Network (GCN), responsible for processing the data. Additionally, in the Graph Attention Networks (GAT), the concept of a "head" is utilized to capture various aspects of the relationships between nodes, with a total of 8 assigned heads. In this approach, we screen 500 stations with an annual demand surpassing 1000. The input features matrix is structured in three dimensions (Batch, Nodes, Features), encompassing variables such as precipitation, pressure, temperature, wind speed, accessibility to population, accessibility to employment, and the in-trips and out-trips from the last time step. We incorporate two layers of graph attention, setting the batch size at 64, the ReLU activation function was employed, the loss function utilized was the mean squared error, and the optimizer used was Adam with a learning rate set to 0.001. The model considers the graph as dynamic, and the total input adjacency matrix is sized at 8,768. The structure of the input matrix, which represents the features of each node, is three-dimensional, with dimensions denoted as (Batch, Nodes, Features). Following training, the mean squared error (MSE) is 0.39, the root mean squared error (RMSE) is 0.63, and the R-squared value is 0.68.

\textbf{Multi Attention Based Graph Convolution model(MGAT): }The objective of the proposed architecture is to enhance the network's performance by creating distinct graphs separately. We have developed three types of graphs, with each component comprising two layers of Graph Attention Networks, each assigned eight heads. This architecture encodes the graph, which is subsequently decoded by fully connected networks. In this methodology, we evaluate 500 stations exhibiting an annual demand exceeding 1000. The input features matrix is organized into three dimensions (Batch, Nodes, Features), incorporating factors like precipitation, pressure, temperature, wind speed, accessibility to population, accessibility to employment, and the in-trips and out-trips from the last time step. The batch size is set at 64, and the ReLU activation function was employed. The loss function utilized was the mean squared error, and the optimizer used was Adam, with a learning rate set to 0.001. and the graph is considered dynamic. We filter stations with an annual travel demand exceeding 1000. The total input adjacency matrix for each component is sized at 8,768. Following training, the Mean Squared Error (MSE) is 0.38, the Root Mean Squared Error (RMSE) is 0.62, and the R-squared value is 0.79.

\textbf{Proposed model:} 
In this study, the travel demand for each station was recorded using a time step of one hour, thereby establishing the structure of a 3-dimensional matrix (Batch, Nodes, Features). A total of 500 stations were selected, all of which experienced demand exceeding 1000 within a year. The model implementation encompasses 2 gate graph convolution layers, incorporating four gates in the first layer and three gates in the second layer, followed by a fully connected layer. The training process involved 100 epochs, with a batch size of 64. Furthermore, a min-max scaler was employed to normalize the matrix features. The ReLU activation function was employed. The loss function utilized was the mean squared error, and the optimizer used was Adam with a learning rate set to 0.001. The feature matrix incorporates variables such as precipitation, pressure, temperature, wind speed, accessibility to population, accessibility to employment, and the in-trips and out-trips from the last time step. These choices were made to ensure the effectiveness and robustness of the model. Over the course of one year, we generated an adjacency matrix for each hour, resulting in a total of 8,768 adjacency matrices, covering the entire span of 2019.

After training the model on this dataset, we evaluated its performance using various metrics. The results indicate an R-square value of 0.82, and the mean squared error (MSE) is calculated to be 0.37; furthermore, the root mean squared error (RMSE) is measured at 0.61.

\autoref{tab:various model} provides a summary and illustration of the results obtained from various classical and machine learning models with their corresponding attributes. Based on the findings presented in \autoref{tab:various model}, it can be concluded that the proposed model demonstrates superior performance compared to the other models. A comprehensive analysis of the $R^2$ value in the main model reveals its superior performance compared to other models. However, it is important to note that the proposed model exhibits slightly higher values of RMSE and MSE when compared to some base models. This disparity can be attributed to the utilization of different sample sizes for each model,specific modeling techniques used, and data distribution, thereby influencing the comparative results. Some classical and machine learning models are unable to handle large volumes of data effectively. In other words, if a large dataset is used as input for these models, their performance tends to decrease. Additionally, this investigation employed a variety of attributes for each model in order to enhance their performance. Therefore, while the proposed model may have higher errors, it still offers improved performance overall.

\begin{figure}
     \centering
     \begin{subfigure}[!ht]{0.3\textwidth}
         \centering
         \includegraphics[width=0.9\linewidth, height=3cm]{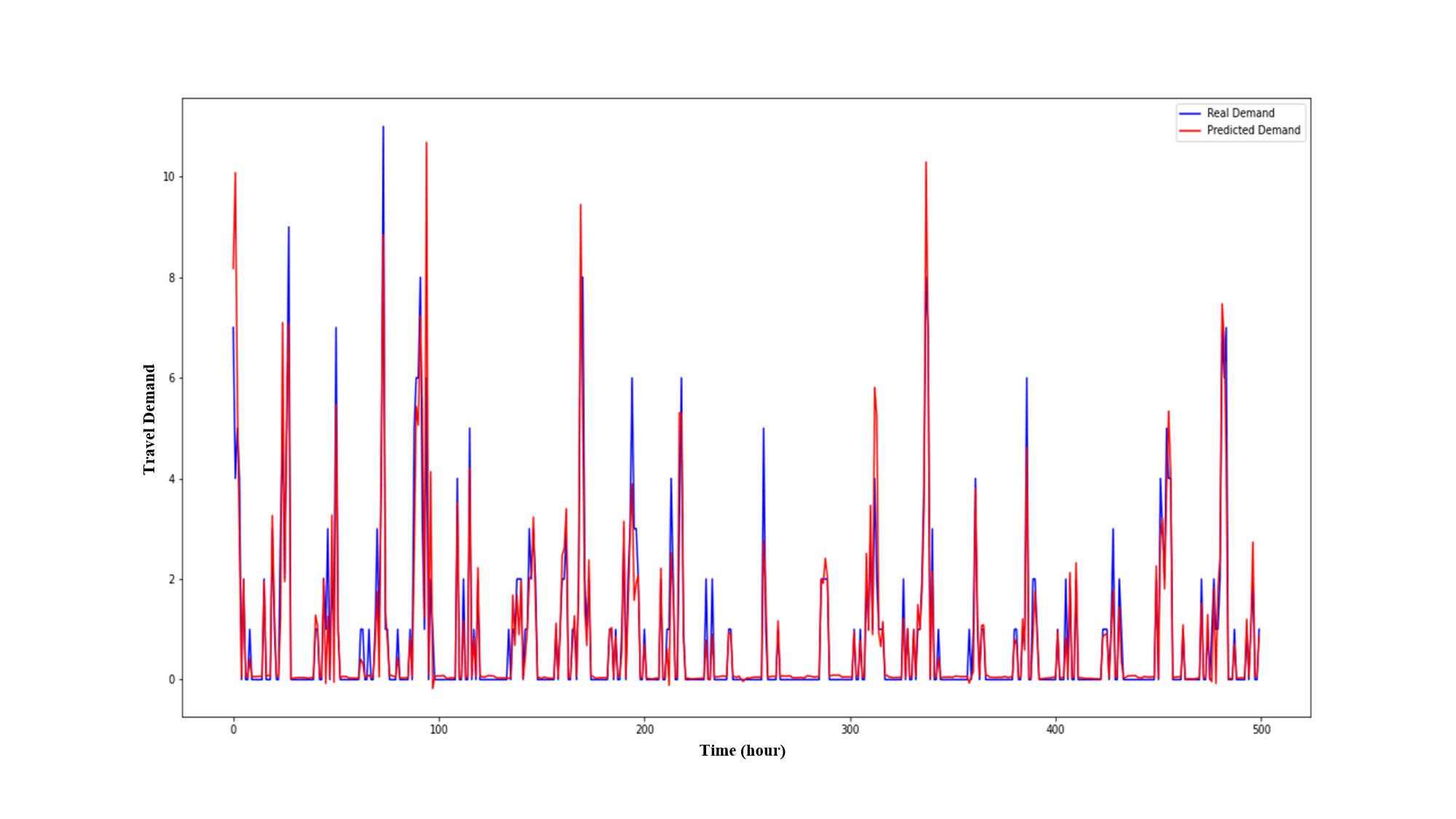}
         \caption{Displays the predicted and actual values of the travel demand for station ID 41, achieving an $R^2$ value of 85\%}
         \label{fig:41_85}
     \end{subfigure}
      \hspace*{3mm}
     \begin{subfigure}[!ht]{0.3\textwidth}
         \centering
         \includegraphics[width=0.9\linewidth, height=3cm]{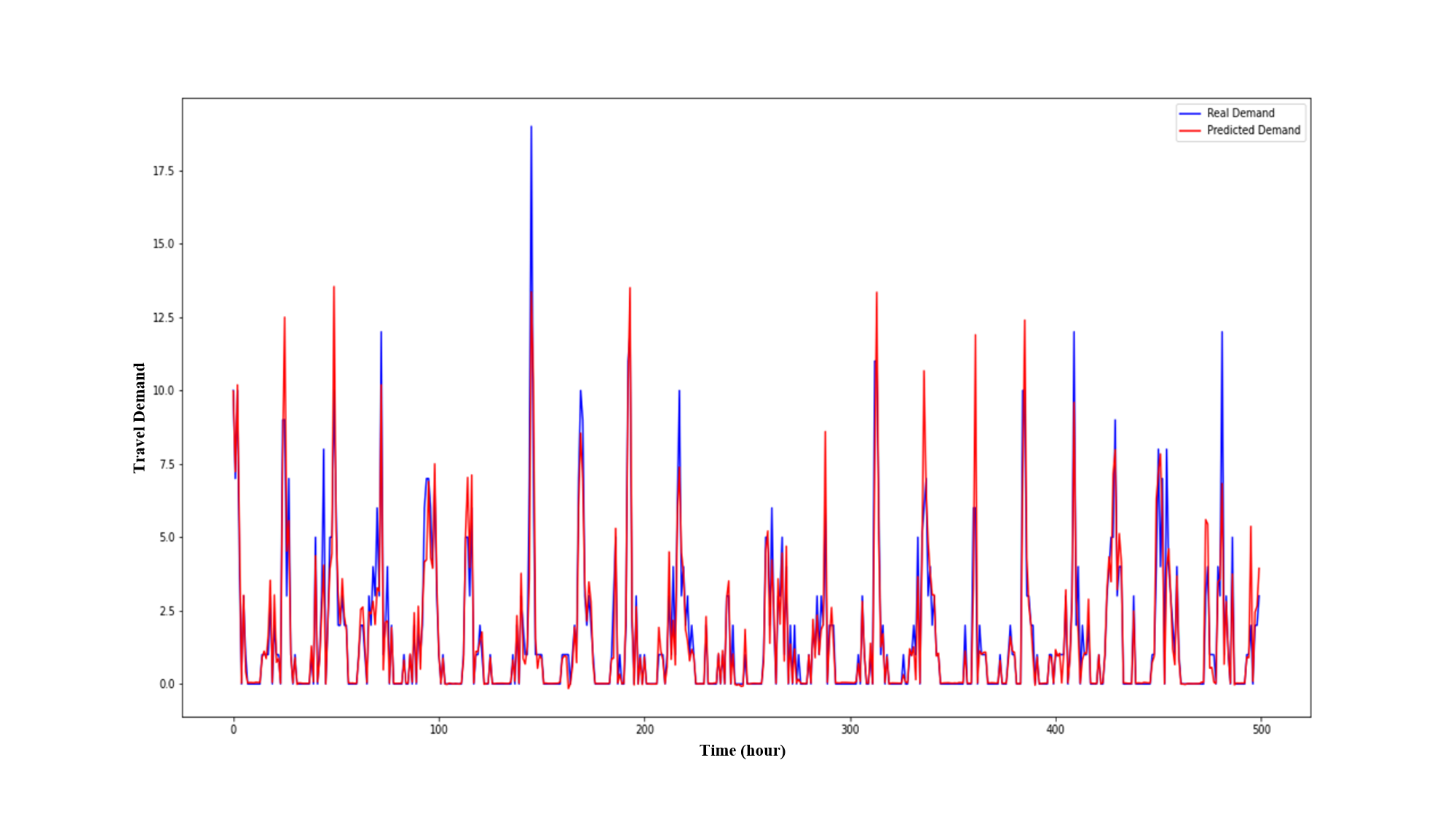}
         \caption{Displays the predicted and actual values of the travel demand for station ID 54, achieving an $R^2$ value of 86\%}
         \label{fig:54_86}
     \end{subfigure}
     \\
     \begin{subfigure}[!ht]{0.3\textwidth}
         \centering
         \includegraphics[width=0.9\linewidth, height=3cm]{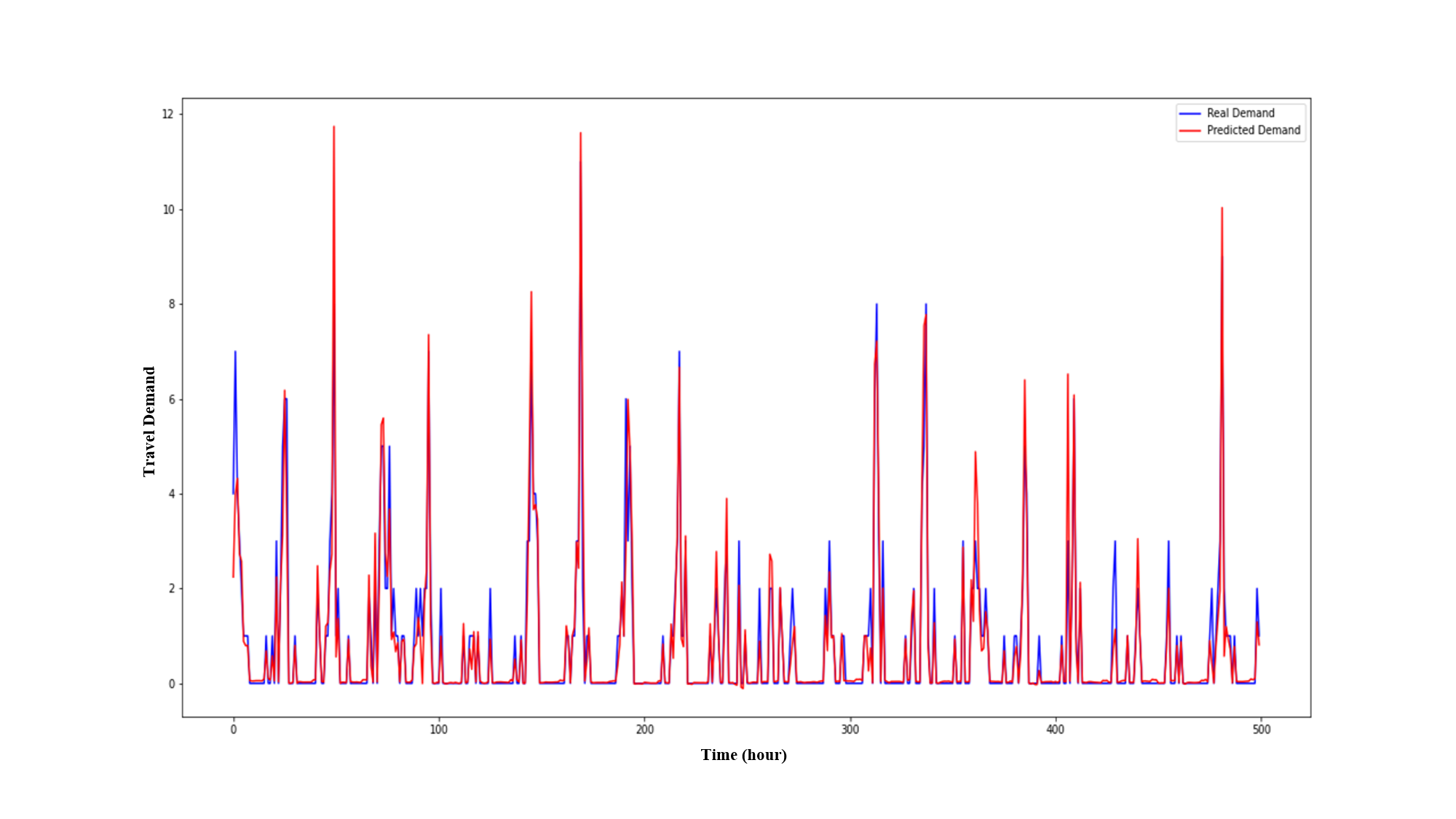}
         \caption{Displays the predicted and actual values of the travel demand for station ID 70, achieving an $R^2$ value of 87\%}
         \label{fig:70_87}
     \end{subfigure}
      \hspace*{3mm}
     \begin{subfigure}[!ht]{0.3\textwidth}
         \centering
         \includegraphics[width=0.9\linewidth, height=3cm]{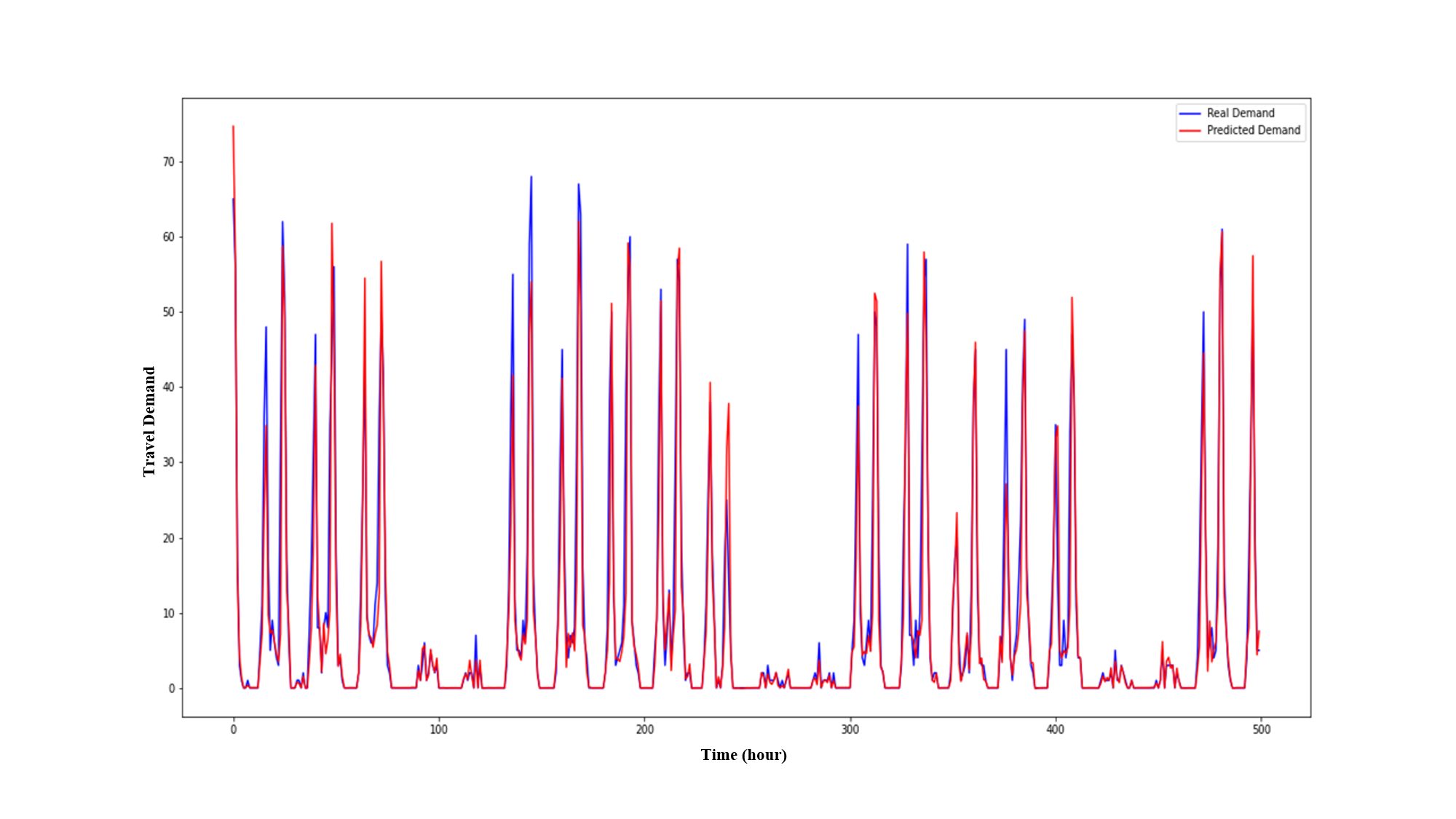}
         \caption{Displays the predicted and actual values of the travel demand for station ID 76, achieving an $R^2$ value of 85\%}
         \label{fig:76_85}
      \end{subfigure}
     \\
     \begin{subfigure}[!ht]{0.3\textwidth}
         \centering
         \includegraphics[width=0.9\linewidth, height=3cm]{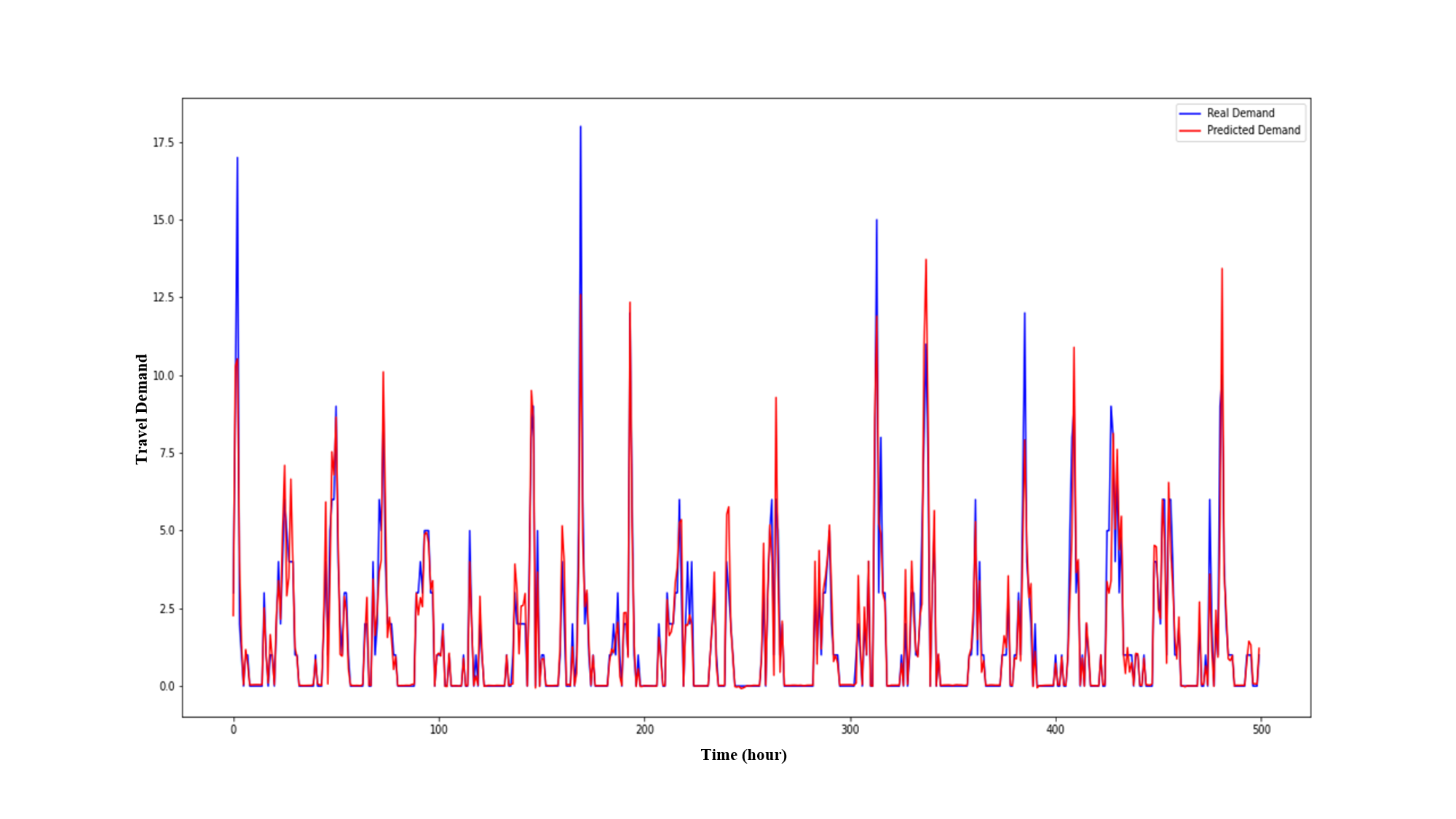}
         \caption{Displays the predicted and actual values of the travel demand for station ID 78, achieving an $R^2$ value of 85\%}
         \label{fig:78_85}
     \end{subfigure}
      \hspace*{3mm}
     \begin{subfigure}[!ht]{0.3\textwidth}
         \centering
         \includegraphics[width=0.9\linewidth, height=3cm]{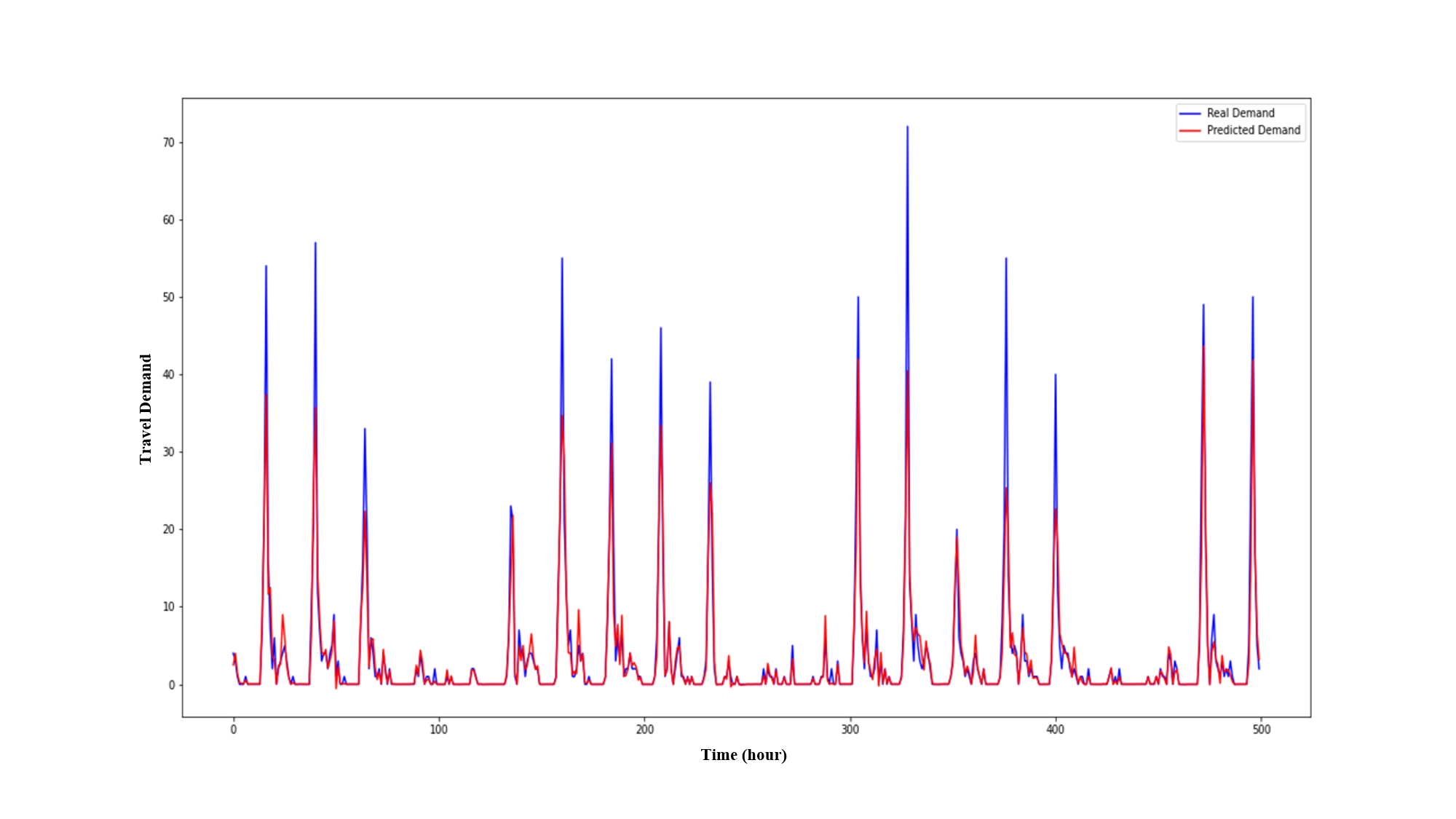}
         \caption{Displays the predicted and actual values of the travel demand for station ID 84, achieving an $R^2$ value of 84\%}
         \label{fig:84_84}
                  
     \end{subfigure}
      \\
     \begin{subfigure}[!ht]{0.3\textwidth}
         \centering
         \includegraphics[width=0.9\linewidth, height=3cm]{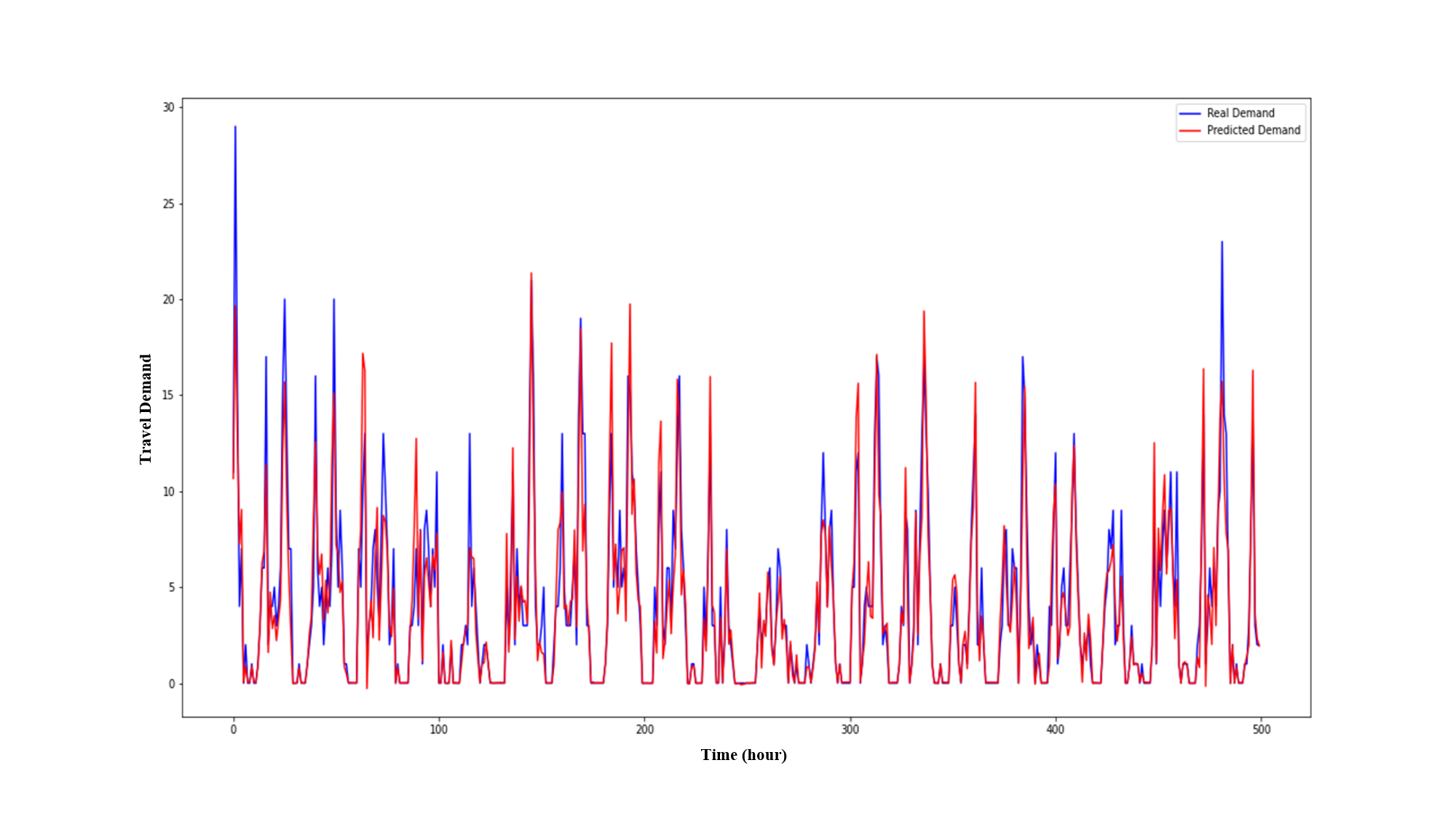}
         \caption{Displays the predicted and actual values of the travel demand for station ID 91, achieving an $R^2$ value of 85\%}
         \label{fig:91_85}
     \end{subfigure}
      \hspace*{3mm}
     \begin{subfigure}[!ht]{0.3\textwidth}
         \centering
         \includegraphics[width=0.9\linewidth, height=3cm]{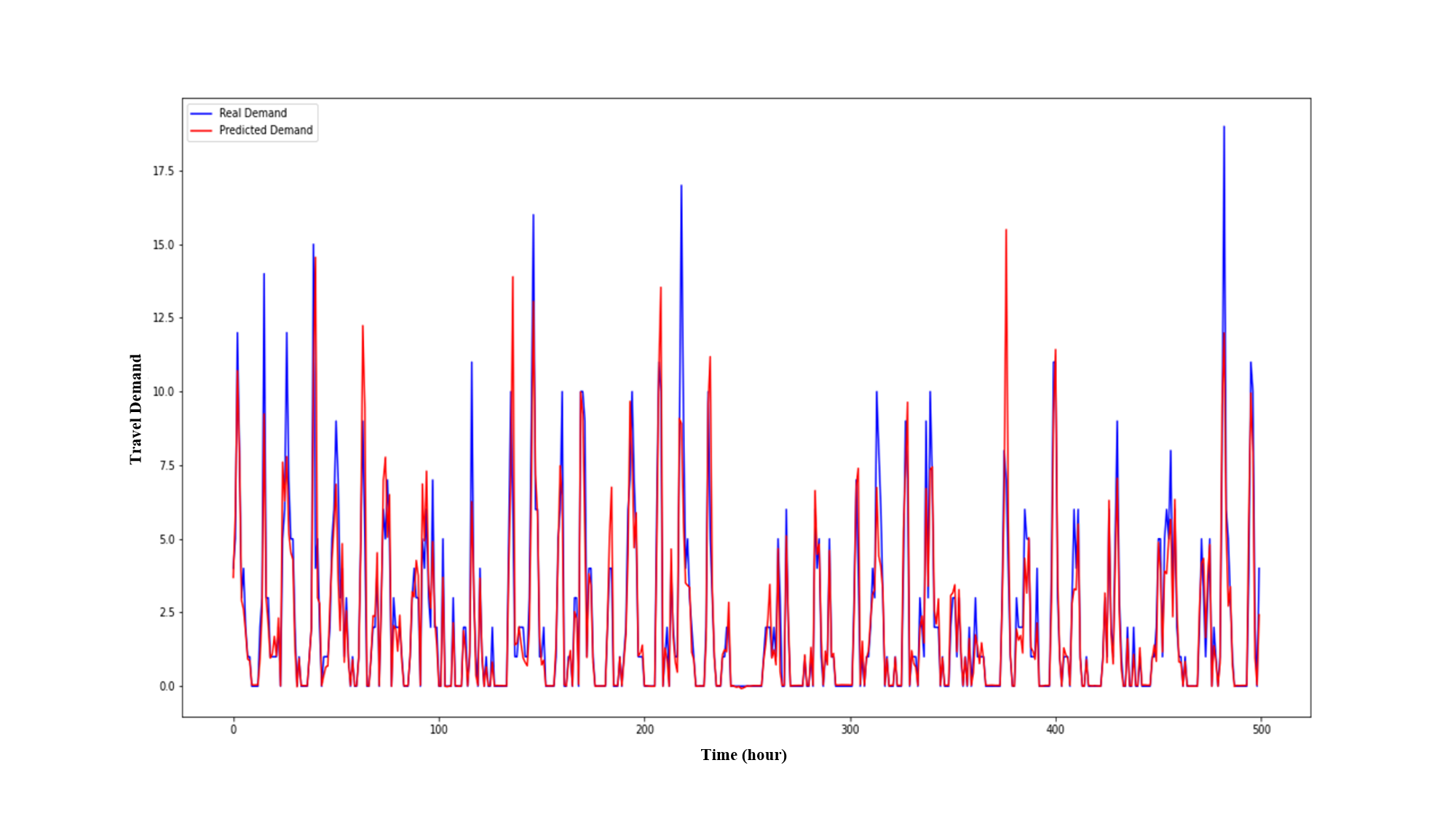}
         \caption{Displays the predicted and actual values of the travel demand for station ID 96, achieving an $R^2$ value of 79\%}
         \label{fig:96_79}
     \end{subfigure}
      \\
     \begin{subfigure}[!ht]{0.3\textwidth}
         \centering
         \includegraphics[width=0.9\linewidth, height=3cm]{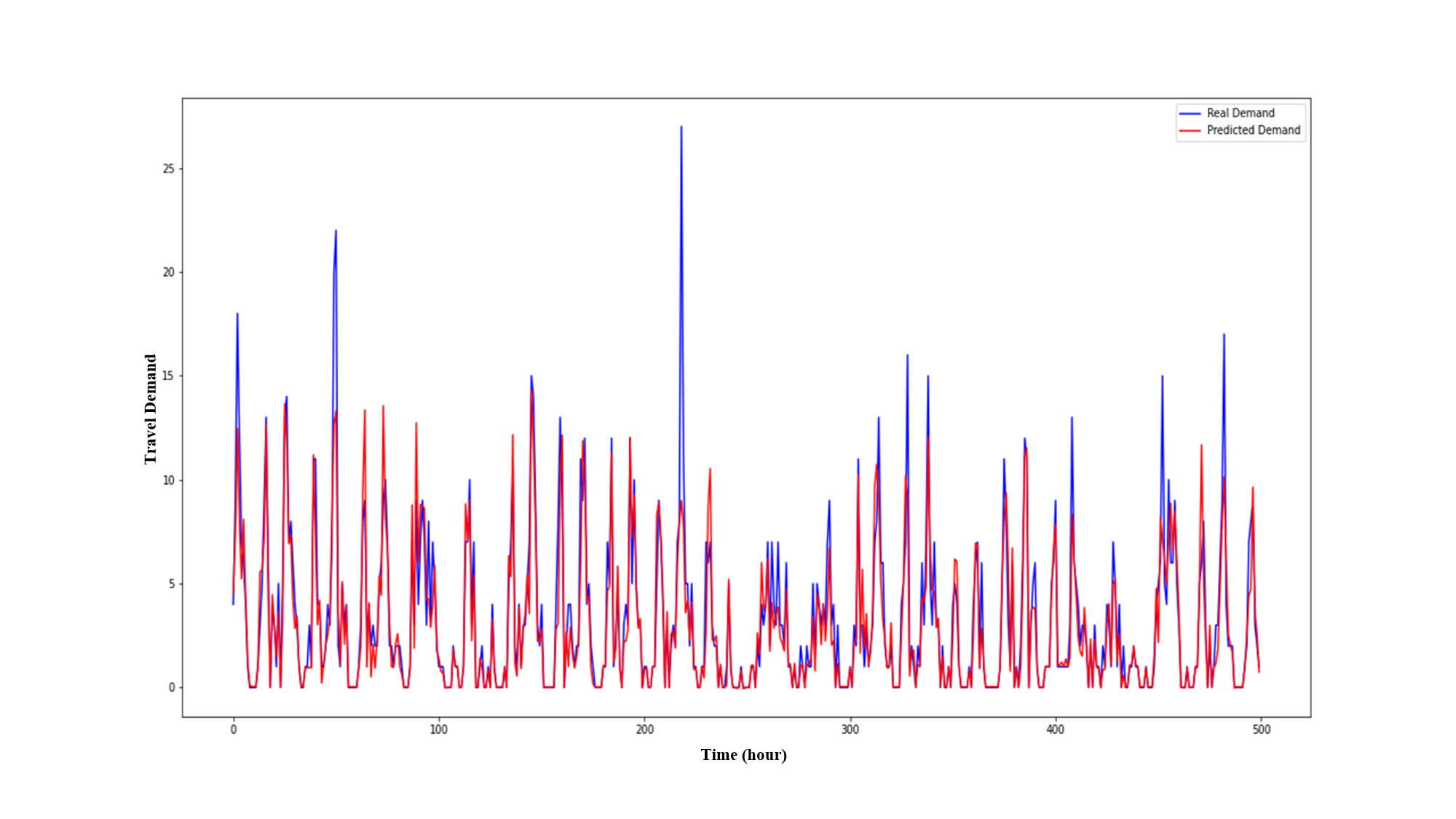}
         \caption{Displays the predicted and actual values of the travel demand for station ID 98, achieving an $R^2$ value of 84\%}
         \label{fig:98_84}
     \end{subfigure}
      \hspace*{3mm}
     \begin{subfigure}[!ht]{0.3\textwidth}
         \centering
         \includegraphics[width=0.9\linewidth, height=3cm]{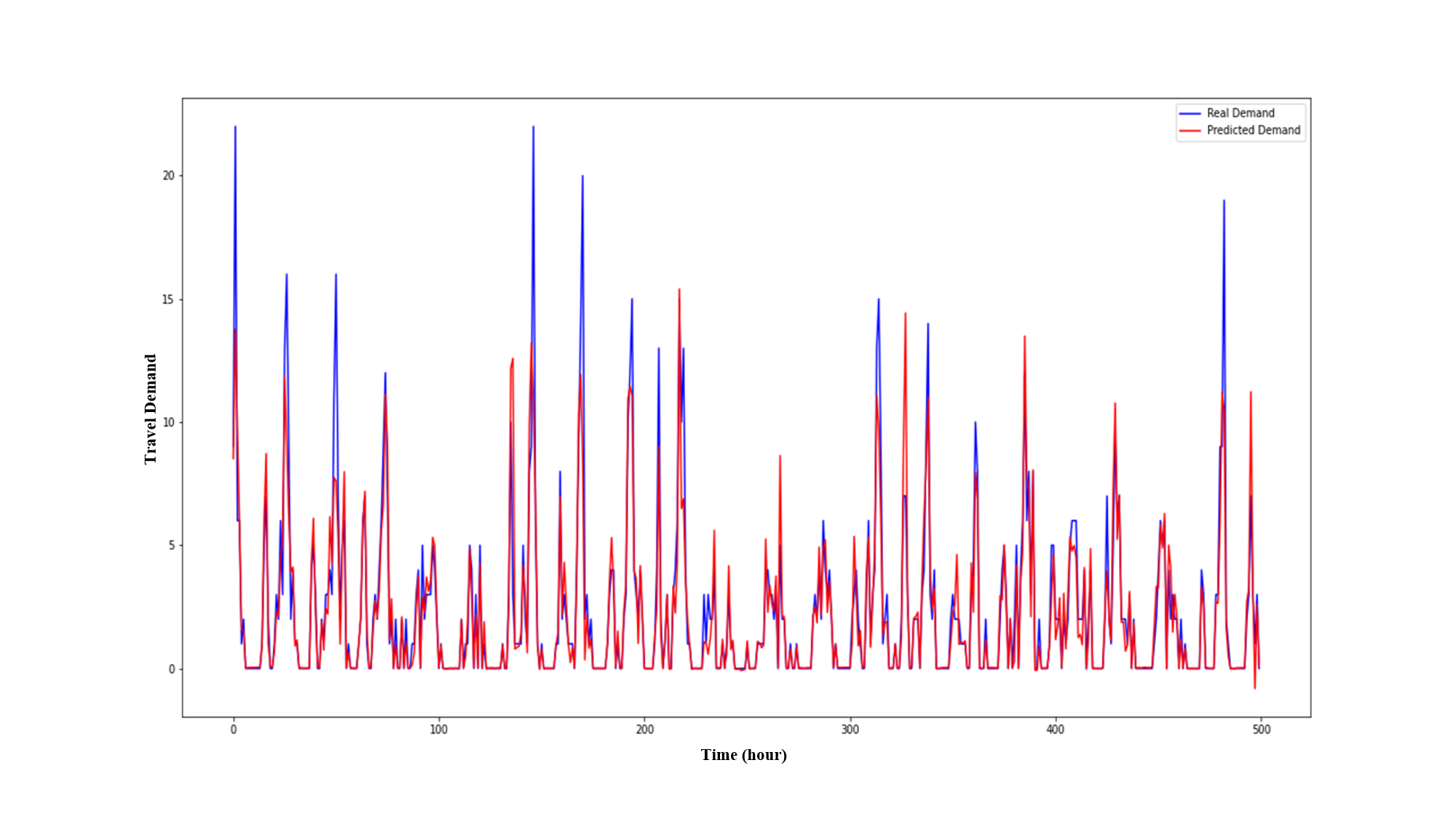}
         \caption{Displays the predicted and actual values of the travel demand for station ID 99, achieving an $R^2$ value of 84\%}
         \label{fig:99_84}
                  
     \end{subfigure}

     \caption{The historical forecast outcomes for each station by the proposed model.}
 \label{fig:prediction}
 \end{figure}

\autoref{fig:prediction} presents a blue plot illustrating the real travel demand during several hours, and the red plot demonstrates the predicted values at various times across multiple stations. The blue plot represents the actual observed travel demand, providing insights into the actual passenger flow over the specified time period. On the other hand, the red plot showcases the predicted values, which are estimated values derived from the prediction of the gate graph convolution neural network. By comparing the blue and red plots, one can analyze the accuracy of the predictions and assess the performance of the model in capturing real travel demand patterns. \autoref{fig:prediction} serves as a visual tool to evaluate the effectiveness of the prediction method and understand any discrepancies or similarities between the predicted and actual travel demand values.

The objective of the investigative task is to accurately predict demand for the next hour. In this endeavor, it is imperative to consider both R-squared and RMSE metrics. The concept underlying the R-squared formula delineates the proportion of variance in the dependent variable (target) that the model predicts from the independent variables (features), essentially gauging how effectively the model elucidates the variability in the data. Given the focus on short-term demand prediction, it is noteworthy that many stations may register zero demand. If a model solely predicts zeros, the RMSE is low. Consequently, relying solely on RMSE may not adequately capture the model's performance. Interestingly, previous investigations often neglected the significance of R-squared values. Nevertheless, an exclusive focus on R-squared for evaluating model performance is unsuitable, as a high R-squared could indicate potential overfitting to the training database. Thus, for such scenarios, a comprehensive evaluation necessitates the simultaneous consideration of both RMSE and R-squared.

\begin{figure}
     \centering
     \begin{subfigure}[!ht]{0.3\textwidth}
         \centering
         \includegraphics[width=0.9\linewidth, height=3cm]{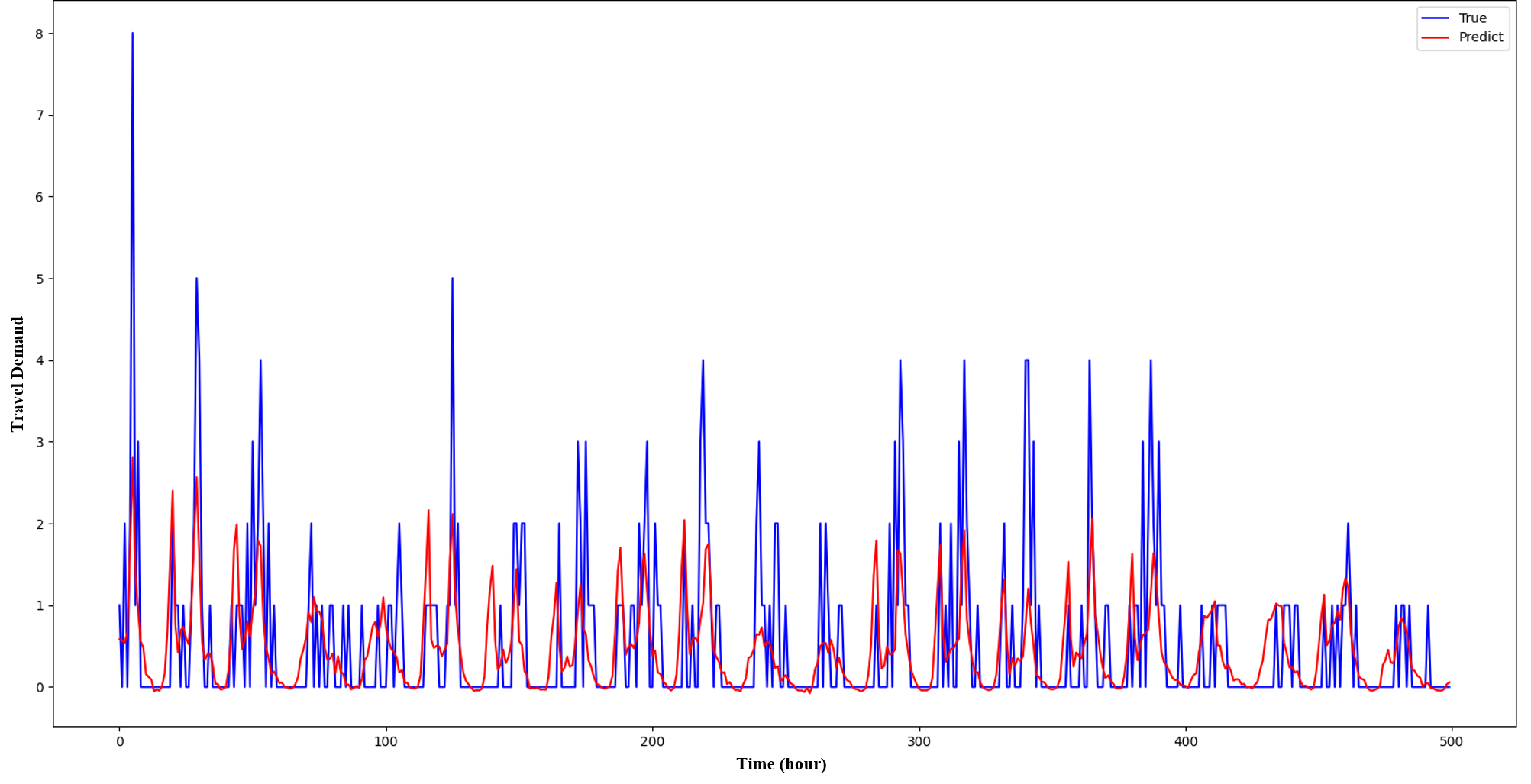}
         \caption{Displays the predicted by GCN-LSTM model and actual values of the travel demand for station ID 163, achieving an $R^2$ value of 32\%}
         \label{fig:163-32-GCNLSTM}
     \end{subfigure}
      \hspace*{3mm}
     \begin{subfigure}[!ht]{0.3\textwidth}
         \centering
         \includegraphics[width=0.9\linewidth, height=3cm]{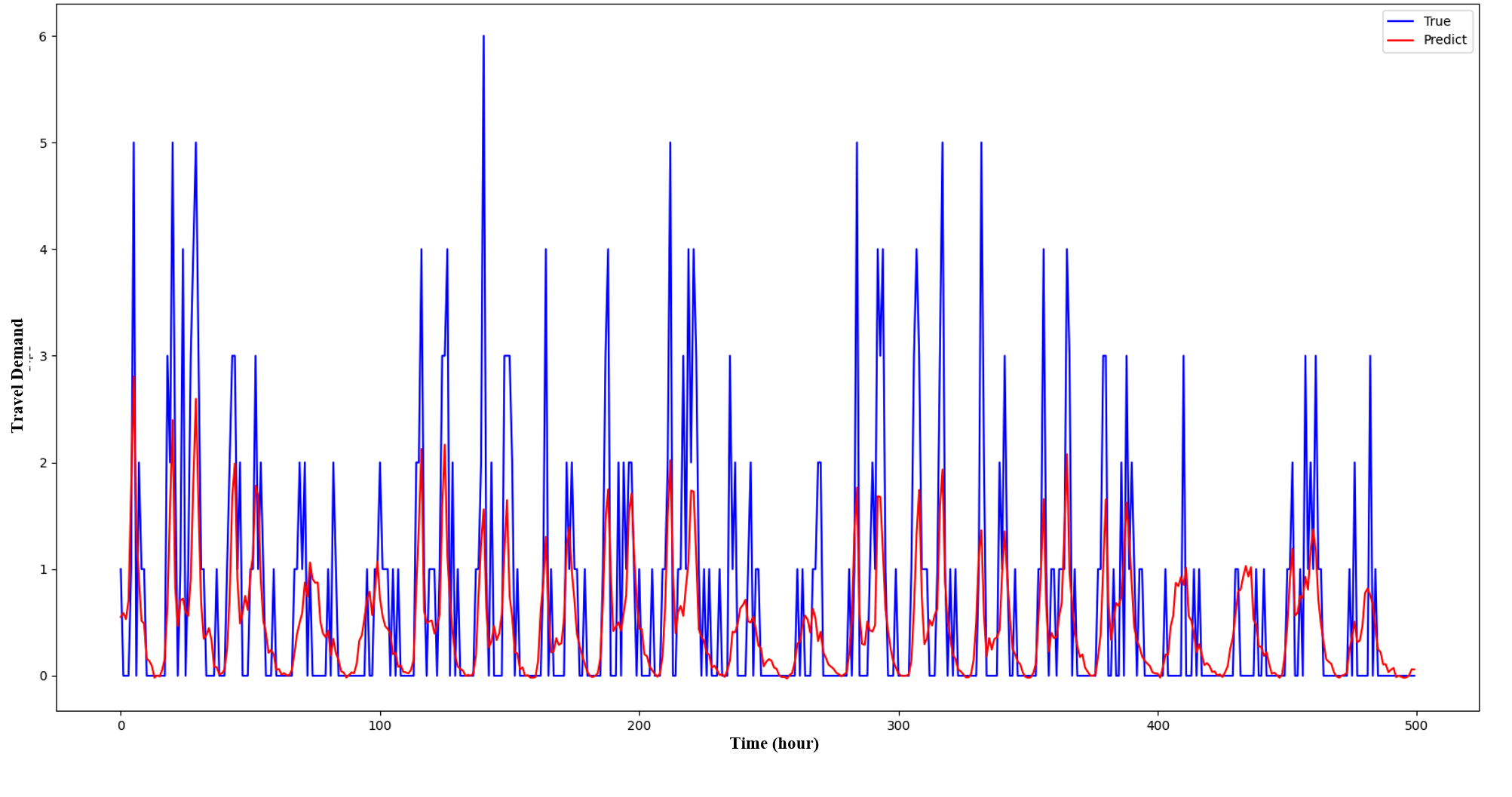}
         \caption{Displays the predicted by GCN-LSTM model and actual values of the travel demand for station ID 315, achieving an $R^2$ value of 42\%}
         \label{fig:315-42-GCNLSTM}
                  
     \end{subfigure}
      \\
     \begin{subfigure}[!ht]{0.3\textwidth}
         \centering
         \includegraphics[width=0.9\linewidth, height=3cm]{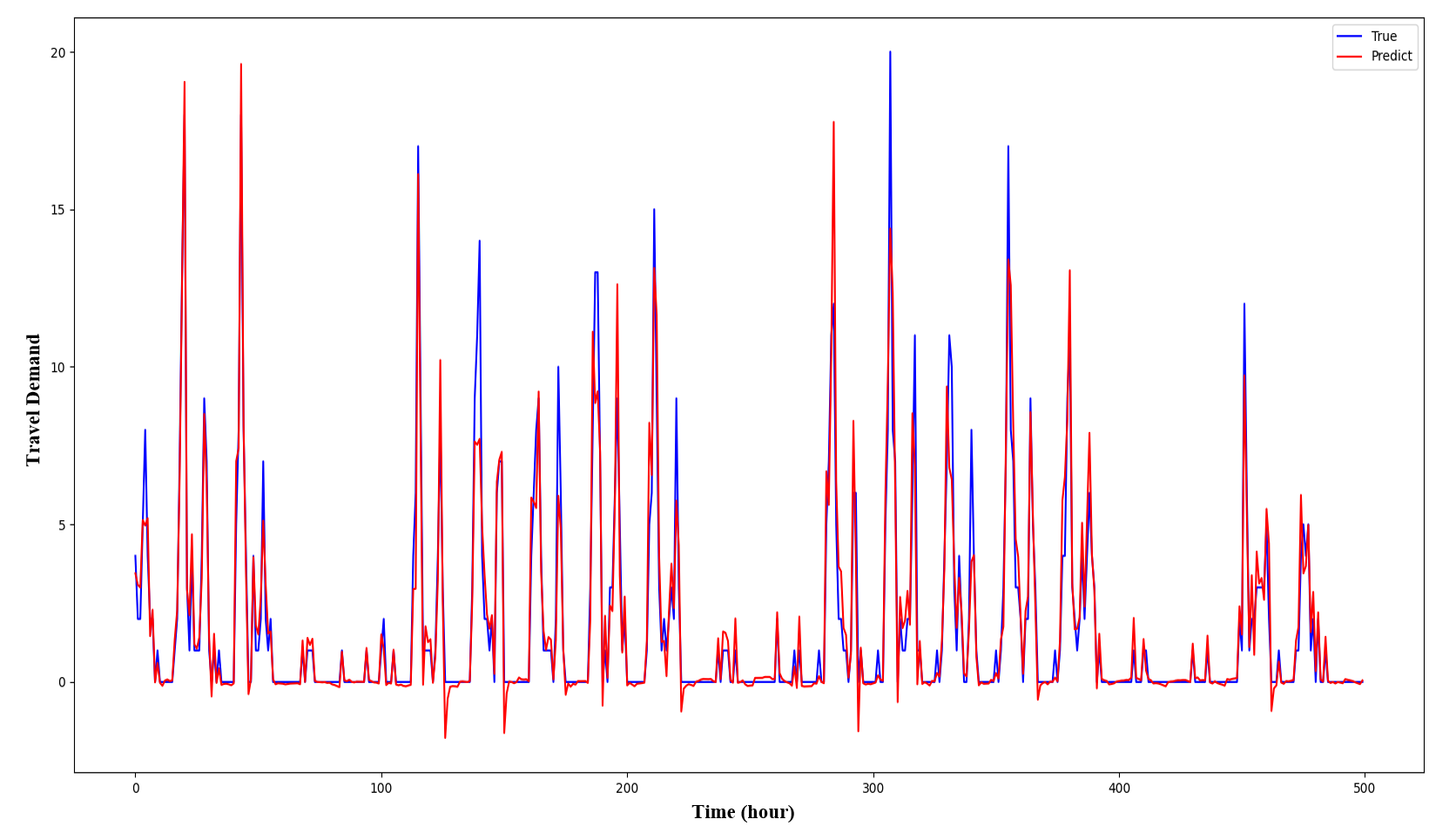}
         \caption{Displays the predicted by ASTGCN model and actual values of the travel demand for station ID 420, achieving an $R^2$ value of 90\%}
         \label{fig:420-90-ASTGCN}
     \end{subfigure}
      \hspace*{3mm}
     \begin{subfigure}[!ht]{0.3\textwidth}
         \centering
         \includegraphics[width=0.9\linewidth, height=3cm]{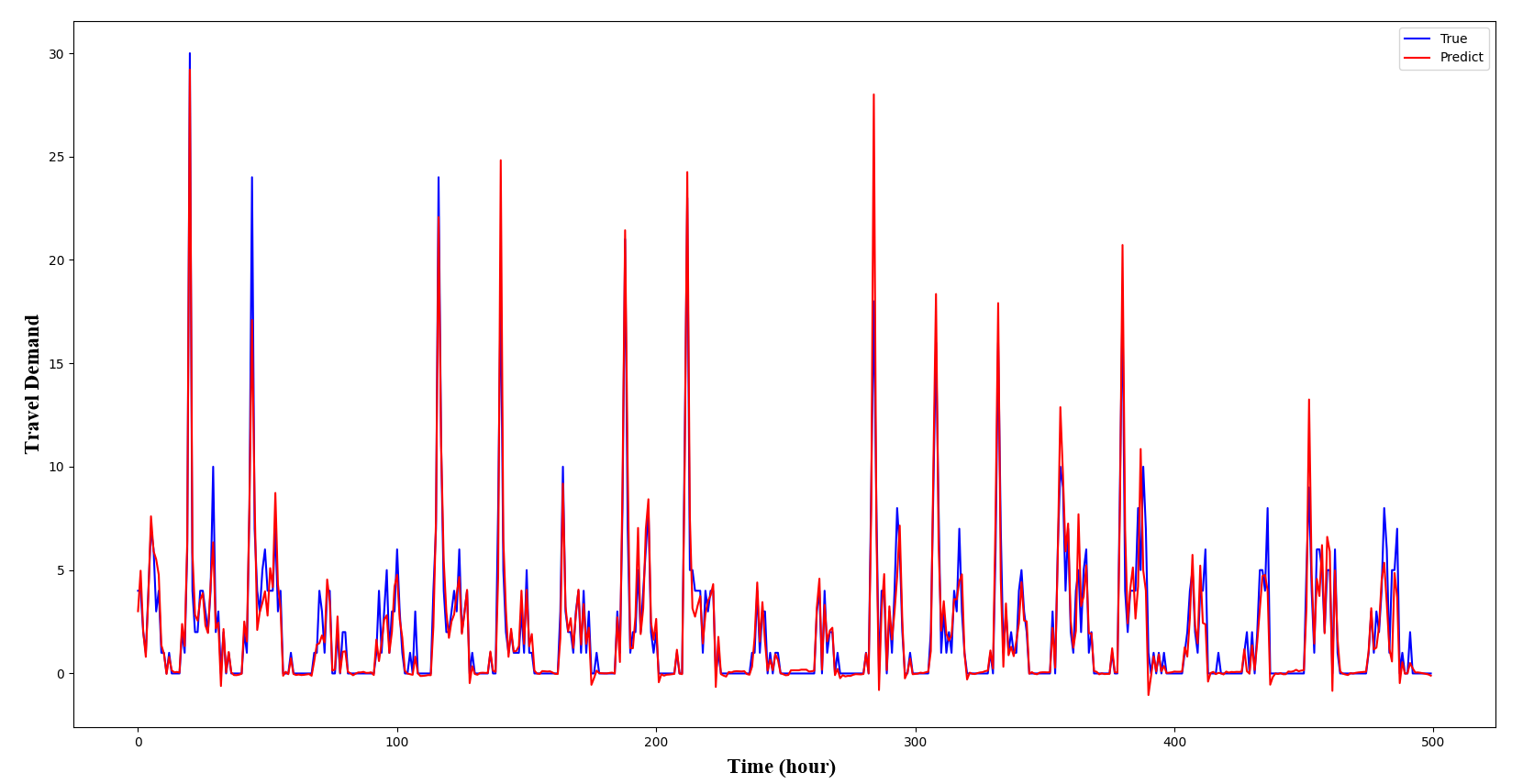}
         \caption{Displays the predicted by ASTGCN model and actual values of the travel demand for station ID 68, achieving an $R^2$ value of 91\%}
         \label{fig:68-91-ASTGCN}
     \end{subfigure}
      \\
     \begin{subfigure}[!ht]{0.3\textwidth}
         \centering
         \includegraphics[width=0.9\linewidth, height=3cm]{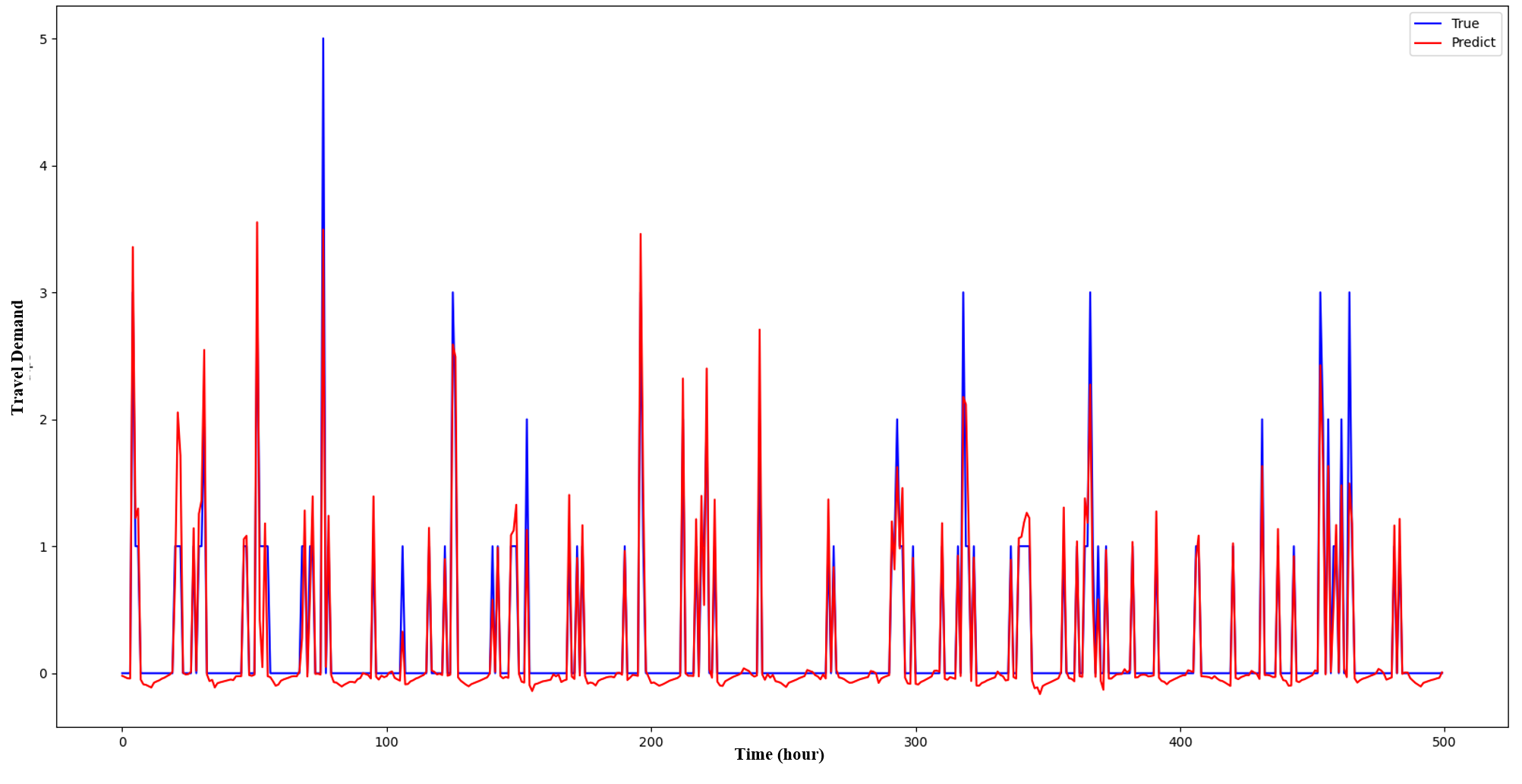}
         \caption{Displays the predicted by MGAT model and actual values of the travel demand for station ID 72, achieving an $R^2$ value of 88\%}
         \label{fig:110-95-MGAT}
     \end{subfigure}
      \hspace*{3mm}
     \begin{subfigure}[!ht]{0.3\textwidth}
         \centering
         \includegraphics[width=0.9\linewidth, height=3cm]{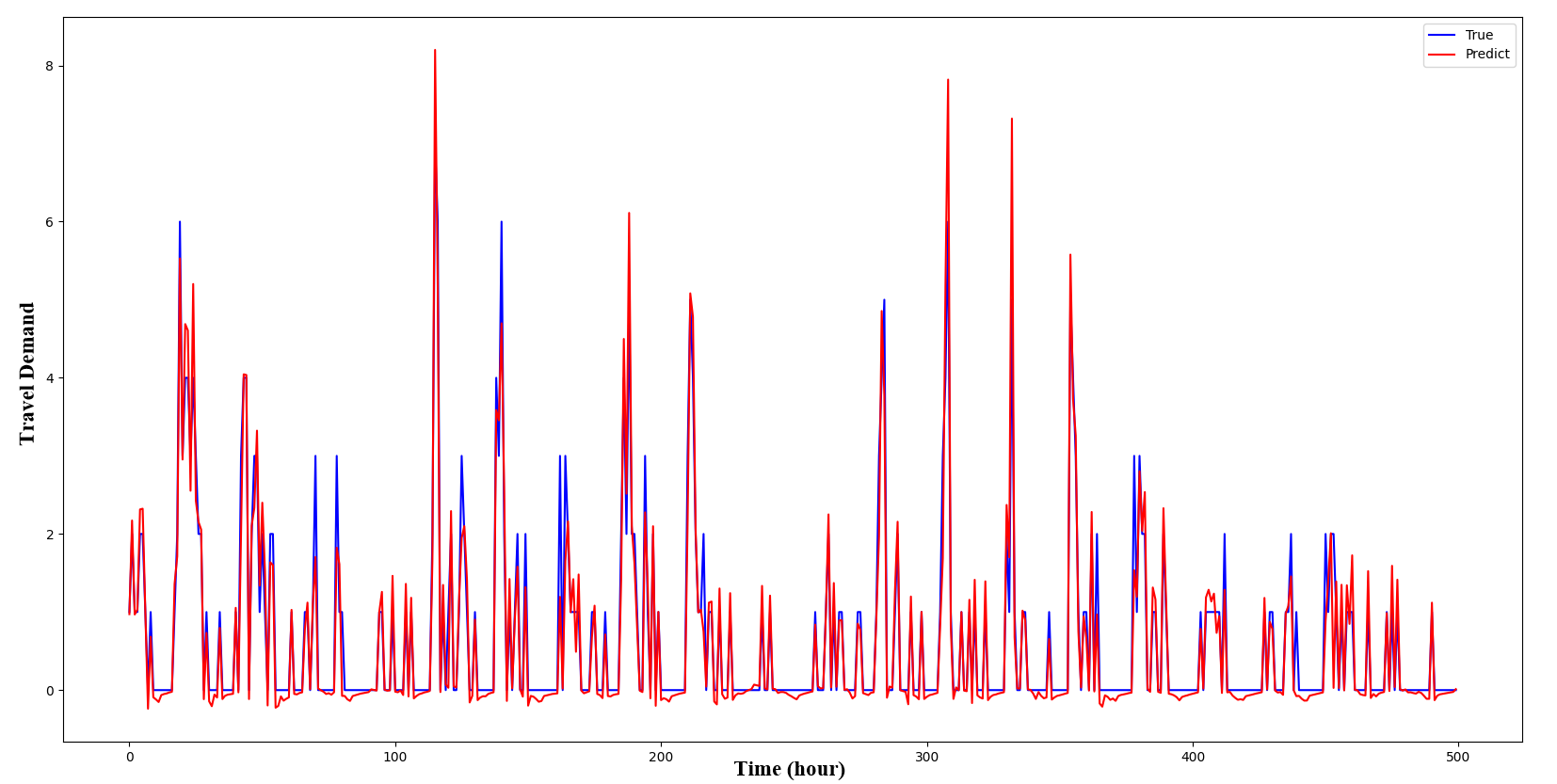}
         \caption{Displays the predicted by MGAT model and actual values of the travel demand for station ID 423, achieving an $R^2$ value of 87\%}
         \label{fig:423-87}
                  
     \end{subfigure}

     \caption{The historical forecast outcomes for each station by benchmark models.}
 \label{fig:prediction-benchmark models}
 \end{figure}

As illustrated in \autoref{fig:prediction-benchmark models}, several samples depict the predicted and actual travel demand, categorized by types of GCN networks. The GCN-LSTM and GAT-LSTM model can discern the trend of actual demand; however, its adherence to actual values is lower than that of other GCN models shown in \autoref{fig:prediction-benchmark models} (a) and \autoref{fig:prediction-benchmark models} (b). The practical advantage of the attention mechanism lies in extracting crucial characteristics vital for the forecasting model, as demonstrated in \autoref{fig:prediction-benchmark models} (c) and \autoref{fig:prediction-benchmark models} (d), and also detailed in \autoref{tab:various model}, where graph attention networks exhibit superior performance compared to GCN-LSTM.

Ultimately, the MGAT model, as presented in \autoref{fig:prediction-benchmark models} (e) and \autoref{fig:prediction-benchmark models} (f), exhibits performance closely comparable to our model. Nevertheless, the primary emphasis of our model is on preserving essential features during training and averting the vanishing gradient. Furthermore, one strategy for addressing the vanishing gradient issue involves training the model with a limited amount of data, such as utilizing a portion of a year's transaction data for training. However, the goal of this investigation is to train the model using the complete dataset for an entire year.

\begin{landscape}
\begin{table}[!ht]
\setlength{\tabcolsep}{10pt} 
\renewcommand{\arraystretch}{1.5} 
    \centering
    \caption{Summary of the performance of various models}
    \begin{tabular}{lcccl}
    \toprule
      Model  & $R^2$  &  $MSE$  & $RMSE$ & Attributes \\
    \midrule
         OLS & 0.5 & 0.49 & 0.7 & precipitation, pressure, temp, wind speed, access to population,\\ &&&&  access to employment, last time step in-trips, out-trips\vspace{0.5em}\\
         ARIMA & 0.1 & 5.6 & 2.37 & Sequence demand \vspace{1em}\\
         MLP & 0.66 & 0.11 & 0.32 & precipitation, pressure, temp, wind speed, access to population,\\ &&&&  access to employment, last time step in-trips, out-trips \vspace{0.5em}\\
         CNN & 0.11 & 2.3 & 1.24 & Sequence demand \vspace{1em}\\
         LSTM & 0.18 & 2.06 & 1.43 & Sequence demand  \vspace{1em}\\
         SVR & 0.32 & 0.68 & 0.82 & precipitation, pressure, temp, wind speed, access to population,\\ &&&&  access to employment, last time step in-trips, out-trips \vspace{0.5em}\\
         GCN & 0.44 & 0.85 & 0.92 & precipitation, pressure, temp, wind speed, access to population ,\\ &&&&   access to employment, last time step in-trips, out-trips \vspace{0.5em}\\
         XGboost & 0.65 & 0.35 & 0.59 & precipitation, pressure, temp, wind speed, access population ,\\ &&&&   access to employment, last time step in-trips, out-trips \vspace{0.5em}\\
         GCN-LSTM & 0.32 & 0.42 & 0.65 & precipitation, pressure, temp, wind speed, access population ,\\ &&&& access to employment , last time step in-trips, out-trips \vspace{0.5em}\\
         ASTGCN & 0.68 & 0.39 & 0.63 & precipitation, pressure, temp, wind speed, access population ,\\ &&&& access to employment , last time step in-trips, out-trips \vspace{0.5em}\\
         MGAT & 0.79 & 0.38 & 0.62 & precipitation, pressure, temp, wind speed, access population ,\\ &&&& access to employment , last time step in-trips, out-trips \vspace{1em}\\
         Proposed model & 0.82 & 0.37 & 0.61 & precipitation, pressure, temp, wind speed, access population ,\\ &&&& access to employment , last time step in-trips, out-trips \vspace{0.5em}\\
         
    \bottomrule
    \end{tabular}
    
    \label{tab:various model}
\end{table}
\end{landscape}

\section{Discussion and conclusion}
\label{sec:discussion}
In summarizing the findings, predicting travel demand in the bike-sharing system encounters significant challenges outlined in the literature section. Notably, as indicated in \autoref{tab:various model}, many traditional, machine learning, and deep learning models, despite showcasing high performance, struggled to predict demand accurately. However, with the introduction of the proposed graph convolution network, it became feasible to consider spatial interactions between nodes, defined in this paper as stations.

A critical observation from \autoref{tab:various model} is that traditional models do not necessarily always exhibit lower performance than machine learning and deep learning models. This point has been widely discussed in various investigations and is reiterated in the context of our study. Specifically, the Ordinary Least Squares (OLS) model demonstrates a higher likelihood of success compared to Support Vector Regression (SVR), Long Short-Term Memory (LSTM), and Convolutional Neural Network (CNN). Additionally, OLS has the advantage of revealing valuable information, such as the significant relationship between dependent and independent variables.

However, our primary objective is to identify a model capable of accurately predicting travel demand. While XGBoost proves to be a powerful method for addressing our problems, the computational demands discussed earlier in the model's conceptualization hinder its support for our extensive dataset.

On the other hand, graph models exhibit superior performance compared to other deep-learning models, as they simultaneously consider spatial and temporal features. It is noteworthy that the Multilayer Perceptron (MLP) model, with only a portion of the database, may outperform some deep-learning models listed in \autoref{tab:various model}.

Our proposed model, designed to retain information through a gate mechanism and prevent the vanishing gradient issue, attains the highest performance among the benchmark models. Additionally, the attention mechanism, as highlighted in \autoref{tab:various model}, exhibits a high potential for accurate predictions. Furthermore, graph convolution networks with gate and attention mechanisms exhibit significant potential in forecasting travel demand. One of the key objectives of this investigation is to implement the gate mechanism in the graph convolution network and assess the model's effectiveness in predicting demand, serving as one of the approaches to address travel demand prediction. As emphasized in the literature review section, various models proposed for training on the historical Divvybikes dataset, such as the TAGCN model, incorporated the attention mechanism. However, the gate mechanism, combined with other successful approaches like the attention mechanism, has also demonstrated the potential to achieve high performance.

The core contribution of this paper is the design of a framework that utilizes a gate graph convolutional neural network for predicting travel demand on an hourly basis, taking into account the dynamic nature of traffic. Through graph convolution with a gate mechanism, we extract spatial and temporal features and train deep-learning models to achieve accurate short-term predictions of bike-sharing traffic flow between stations, resulting in an $R^2$ value that demonstrates an accuracy of $82\%$ and lower RMSE and MSE. When compared to other classic and machine learning models discussed in \autoref{sec:result}, it outperforms them in terms of performance. Furthermore, our findings indicate that the gate graph convolution model outperforms others regarding predictive accuracy and model fit. The results of this study highlight the potential for enhancing travel demand modeling by utilizing deep learning techniques. The integration of trajectory bike-share data, weather data, and population and employment accessibility highlights their complementary nature. This is particularly significant for short-term bicycle demand forecasting, as it relies on factors such as weather conditions (such as temperature, humidity, and wind speed) and the level of access to employment and population centers. Stations with lower accessibility tend to experience fewer trips, while the spatial proximity of stations can also mutually influence each other. This study highlights the importance of incorporating data and employing a spatial and temporal approach to predict traffic flow between stations. By addressing the challenge of predicting travel demand, this approach contributes to the effective rebalancing of bike-sharing services, ultimately leading to improved service quality and increased customer satisfaction.  The experiments illustrate that the framework's capability to forecast short-term traffic flow outperforms popular machine learning, deep learning, and classical techniques. By exploring different models that consider either temporal or spatial approaches, we have concluded that spatio-temporal prediction is the most effective approach. Additionally, the study reveals significant patterns in how urban form influences bike-share traffic behavior across various spatial and temporal contexts.

However, the current model still has potential for further improvement. One way for enhancement is the incorporation of attention mechanisms and heterogeneous graph models to enhance the extraction of spatial features. Additionally, to minimize losses in short-term prediction tasks, future research will focus on exploring more effective frameworks that leverage diverse types of data for traffic flow prediction. Moreover, we will strive to optimize the proposed architecture for extracting spatio-temporal features, ultimately improving the framework's performance and prediction accuracy.
One assumption of the framework is its model generalizability with another bike-share database, as in this investigation, we lack access to alternative databases. We suggest, for future research, utilizing a different database from various cities to train the model, aligning with the proposed framework.

\section{Conflict of interest}
All authors certify that they have no affiliations with or involvement in any organization or entity with any financial interest or non-financial interest in the subject matter or materials discussed in this manuscript.\\
Submitted by\\
Ali Edrisi, Faculty of Civil Engineering, K. N. Toosi University of Technology, Tehran, Iran; (Corresponding author)

\newpage
\bibliographystyle{sn-chicago.bst}
\bibliography{Mylib.bib}

\newpage

\end{document}